\documentclass[10pt,journal,compsoc,table,tikz,x11names,final]{IEEEtran}

\usepackage[pdftex]{graphicx}
\graphicspath{{.}}
\usepackage{pgf}  % render matplotlib plots with bibliography references
\usepackage{epsfig}
\usepackage{xcolor}

\usepackage{amsfonts}
\usepackage{amsmath}
\usepackage{amssymb}
\usepackage{bm}
\usepackage{fontawesome}  % special icons, such as GitHub logo
\usepackage{numprint}  % nicely format big numbers
\usepackage{pifont}

\usepackage{makecell}  % advanced table cell formatting
\usepackage{multirow}  % span table cell from multiple rows
\usepackage{booktabs}  % table formatting (e.g. toprule, midrule, bottomrule)
\usepackage{tabularx}

\usepackage[caption=false]{subfig}
\usepackage{soul}

\usepackage{forest}
\colorlet{linecol}{black!75}
\usepackage{tikz}
\usetikzlibrary{calc,shapes.geometric,trees,positioning,arrows.meta,matrix,decorations.pathreplacing}

\usepackage{enumitem} % additional functionality to enumerations

\definecolor{myred}{HTML}{FC8D59}
\definecolor{myyellow}{HTML}{FFFFBF}
\definecolor{myblue}{HTML}{91BFDB}
\definecolor{lightergray}{rgb}{0.90, 0.90, 0.90}

\DeclareMathOperator*{\argmax}{\arg\!\max}

\newcommand*{\bigrightArrow}{\scalebox{1.5}[2.5]{\ding{223}}}

\makeatletter
\DeclareRobustCommand{\rvdots}{%
  \vbox{
    \baselineskip4\p@\lineskiplimit\z@
    \kern-\p@
    \hbox{.}\hbox{.}\hbox{.}
  }%
}
\makeatother

\tikzset{
    my rounded corners/.append style={rounded corners=2pt},
}
\newcommand{\unit}{0.6cm}

\usepackage{lineno}
\usepackage[pagebackref=true,breaklinks=true,colorlinks,bookmarks=false]{hyperref}
\modulolinenumbers[5]

\usepackage[numbers,sort&compress]{natbib}

\usepackage{flushend} % positions biographies at equal end as references

\begin{document}
    \title{A Review of Emerging Research Directions in Abstract Visual Reasoning}
    \author{
        Miko{\l}aj~Ma{\l}ki{\'n}ski and Jacek~Ma{\'n}dziuk%
        \IEEEcompsocitemizethanks{%
            \IEEEcompsocthanksitem Miko{\l}aj~Ma{\l}ki{\'n}ski is a Ph.D. student at the Doctoral School no. 3, Warsaw University of Technology, pl. Politechniki 1, 00-661 Warsaw, Poland, m.malkinski@mini.pw.edu.pl.%
            \IEEEcompsocthanksitem Jacek~Ma{\'n}dziuk is with the Faculty of Mathematics and Information Science, Warsaw University of Technology, Koszykowa 75, 00-662 Warsaw, Poland, mandziuk@mini.pw.edu.pl.%
        }%
    }

    \IEEEtitleabstractindextext{
        \begin{abstract}
    Abstract Visual Reasoning (AVR) problems are commonly used to approximate human intelligence.
    They test the ability of applying previously gained knowledge, experience and skills in a completely new setting, which makes them particularly well-suited for this task.
    Recently, the AVR problems have become popular as a proxy to study machine intelligence, which has led to emergence of new distinct types of problems and multiple benchmark sets.
    In this work we review this emerging AVR research and propose a taxonomy to categorise the AVR tasks along 5 dimensions: input shapes, hidden rules, target task, cognitive function, and specific challenge.
    The perspective taken in this survey allows to characterise AVR problems with respect to their shared and distinct properties, provides a unified view on the existing approaches to solving AVR tasks, shows how the AVR problems relate to practical applications, and outlines promising directions for future work.
    One of them refers to the observation that in the machine learning literature different tasks are considered in isolation, which is in the stark contrast with the way the AVR tasks are used to measure human intelligence, where multiple types of problems are combined within a single IQ test.
\end{abstract}

    }
    \maketitle
    
    \section{Introduction}\label{sec:introduction}

Abstract Visual Reasoning (AVR) domain encompasses problems that require formulating analogies between abstract visual concepts instantiated in varying scenarios.
Generally, in such tasks the core challenge is to recognise relations that govern simple 2D shapes and their attributes among potentially many images.
These problems oftentimes require the solver to apply previously gained skills, knowledge and experience in a completely new setting and are therefore considered as a well-performing predictor of human intelligence (IQ)~\cite{snow1984topography,carpenter1990one}.
Examples of AVR tasks include the commonly recognized Raven's Progressive Matrices (RPMs)~\cite{raven1936mental,raven1998raven}, Odd-one-out ($\mathrm{O}^3$) tasks~\cite{gardner2006colossal,ruiz2011building} or Bongard Problems (BPs)~\cite{bongard1968recognition}, as well as the emerging ones that focus on extracting and applying visual analogies~\cite{hill2019learning}, combine visual and arithmetic reasoning~\cite{zhang2020machine}, explicitly focus on extrapolation~\cite{webb2020learning}, require the ability to differentiate between same and different concepts~\cite{fleuret2011comparing}, or test generative modelling in few-shot learning setting~\cite{chollet2019measure}.

On the one hand, a recent stream of research has revealed that current machine learning (ML) methods, especially those that fall into the deep learning (DL) bucket, can solve certain AVR problems posed in simpler settings exceptionally well~\cite{hoshen2017iq,mandziuk2019deepiq}.
Other works, however, have pointed out that AVR tasks offer a rich and challenging testbed for evaluating different generalisation capabilities of the tested ML/DL methods~\cite{barrett2018measuring}.
Many of these generalisation challenges remain unsolved to date.
Furthermore, various AVR benchmarks have recently been proposed (e.g.~\cite{hill2019learning,zhang2020machine,nie2020bongard}) that constantly uncover consecutive shortcomings of modern ML methods.

\begin{figure}
  \centering
  \begin{forest}
    for tree={
      line width=1pt,
      grow=east,
      anchor=south,
      parent anchor=south east,
      child anchor=south west,
      align=center,
      reversed=true,
      l sep+=2.5pt,
      s sep+=-2.5pt,
      inner sep=2.5pt,
      outer sep=0pt,
      edge path={
        \noexpand\path [draw, rounded corners=5pt, \forestoption{edge}] (!u.parent anchor) to (.child anchor)\forestoption{edge label} -- (.south east);
      },
      for root={
        my rounded corners,
        draw,
        parent anchor=east,
      },
    }
    [Abstract\\Visual\\Reasoning
      [Input\\shapes
        [Geometric]
        [Abstract]]
      [Hidden\\rules
        [Explicit]
        [Abstract]]
      [Target\\task
        [Classification]
        [Generation]
        [Description]]
      [Cognitive\\function
        [Completion]
        [Discrimination]]
      [Specific\\challenge
        [Domain transfer]
        [Extrapolation]
        [Arithmetic]]]
  \end{forest}
  \caption{%
  \textbf{AVR taxonomy.}
  The discussed AVR tasks can be cataloged along 5 dimensions:
  1) input shapes present in panels of the problem instance;
  2) types of hidden rules that govern shapes and their attributes;
  3) a target task being solved;
  4) a cognitive function imitated when solving the problem;
  5) a specific challenge posed by the task.}
  \label{fig:avr-taxonomy}
\end{figure}
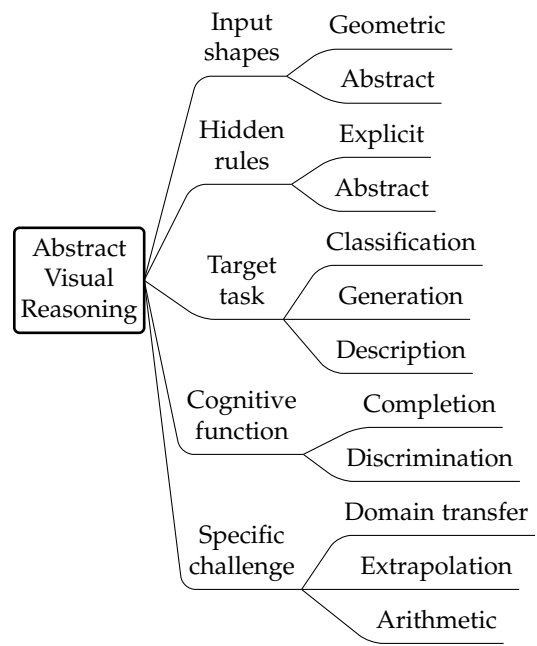

\begin{figure*}
    \centering
    \begin{forest}
        for tree={
            line width=1pt,
            my rounded corners,
            draw=linecol,
            edge={color=linecol, >={Triangle[]}, ->},
            edge path={
                \noexpand\path [draw, rounded corners=5pt, \forestoption{edge}] (!u.parent anchor) to (.child anchor);
            },
            if level=0{%
                l sep+=1.5cm,
                parent anchor=south,
            }{%
                if level=1{%
                    child anchor=north,
                    align=center,
                }{}%
            },
        }
[Visual Reasoning (VR)
    [Abstract Visual \\ Reasoning (AVR) \\ (\cite{hernandez2016computer,mitchell2021abstraction,van2021much,stabinger2021evaluating,malkinski2022deep} and this work)]
    [Visual Question \\ Answering (VQA) \\ \cite{antol2015vqa,wu2017visual,kafle2017visual,manmadhan2020visual,srivastava2020visual}]
    [Visual Commonsense \\ Reasoning (VCR) \\ \cite{yu2016modeling,kahou2017figureqa,suhr2019corpus,zellers2019recognition,xie2019visual}]
    [Physical \\ Reasoning (PR) \\ \cite{santoro2017simple,bakhtin2019phyre,allen2020rapid,riochet2020intphys,baradel2020cophy}]]
    \end{forest}
    \caption{\textbf{AVR and related domains.} The field of AVR includes problems that require reasoning about abstract concepts present in the image or identifying hidden rules that govern the visual entities. Although such concepts are also present in related domains of VQA, VCR and PR, the AVR field differs from them mainly by neither utilizing natural language-based input (as in VQA and VCR) nor referring to the real-world objects (as in VCR and PR).}
    \label{fig:taxonomy}
\end{figure*}
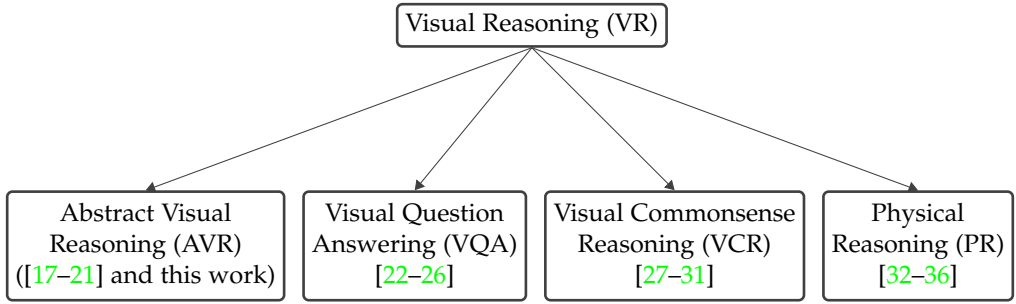

With the increased number of available AVR benchmarks, more and more methods are proposed to tackle them.
Even though in many cases the methods operate on similar inputs and outputs, most of the time they are evaluated only on a single chosen task, without considering the large set of remaining available problems.
As a consequence, highly specialised models that excel in one task may be over-fitted to a particular problem and struggle with even slightly different tasks (e.g. CoPINet~\cite{zhang2019learning} was able to solve matrices from RAVEN~\cite{zhang2019raven} almost perfectly, but was later shown to achieve much worse results on I-RAVEN~\cite{hu2021stratified}).

\subsection{Motivation}
This review collates existing AVR problems in 
%terms of 
several dimensions, which can be used by ML practitioners to make an informed decision when selecting a benchmark for evaluating selected generalisation capabilities of the tested approaches.

By pointing out the commonalities and differences between AVR tasks, we invite the AVR community to consider multiple AVR problems at once when working on new approaches.
Such a setup better aligns with how AVR tasks are used to measure human intelligence and is still a grand challenge for Artificial Intelligence (AI) and ML systems.

On a general note, research on universal multi-task approaches to solving diverse AVR problems may help understand the operational principles of human intelligence and may be a step towards accomplishing Artificial General Intelligence (AGI).

\subsection{Taxonomy}
To better describe and explain the links and differences among a variety of AVR tasks, we propose the following taxonomy that allows to explore and catalogue the AVR problems along 5 dimensions (Fig.~\ref{fig:avr-taxonomy}):
\begin{enumerate}[nosep,leftmargin=*,label=\arabic*)]
    \item \textbf{Input shapes.}
    AVR problems are composed of multiple panels, each of them potentially containing multiple shapes.
    The objects can be of different forms, including \textit{geometric} shapes drawn from a limited vocabulary, and \textit{abstract} shapes that come from a wide set of possibilities and rarely repeat across problems.
    \item \textbf{Hidden rules.}
    Similarly to the input shapes, AVR tasks may comprise \textit{explicit} rules known beforehand that are sampled from a fixed vocabulary.
    Such rules most often refer to logical operators, such as OR (disjunction), AND (conjunction), XOR (exclusive disjunction), etc., combined with other well defined concepts, such as \textit{progression} or \textit{constancy}.
    On the other hand, AVR tasks may contain a practically unlimited set of \textit{abstract} rules defined loosely by means of intuitive visual concepts or expressed in natural language.
    \item \textbf{Target task.}
    AVR problems can be designed with various target tasks in mind.
    In the vast majority of the cases these are:
    \textit{classification}, where the goal is to select an answer from a limited set of choices;
    \textit{generation}, where a missing image (or part of it) has to be generated;
    \textit{description}, where the answer has to be described in natural language.
    \item \textbf{Cognitive function.}
    We further categorise the AVR problems based on the cognitive function that is required to solve them.
    This division includes problems that focus on:
    \textit{completion}, where the goal is to fill-in the problem matrix with an answer;
    \textit{discrimination}, where a rule (or a set of rules) has to be discovered that separates the provided panels.
    \item \textbf{Specific challenge.}
    The last perspective we identify is the type of challenge that underlines a given AVR problem.
    This involves:
    \textit{domain transfer}, where a concept learned in a source domain has to be transferred to a target domain;
    \textit{extrapolation}, that requires solving problems with novel values of certain attributes, not observed in the learning phase;
    \textit{arithmetic}, which tests the ability of reasoning about numbers and arithmetic operations from visual inputs.
\end{enumerate}

\subsection{Scope}\label{subsec:scope}
In a broader perspective, the field of AVR belongs to the Visual Reasoning domain which encompasses tasks that require reasoning, i.e. identifying and extracting task-relevant information from visual input (usually an image or a set of images).
%After recent renaissance of applying DL to visual problems, which started with impressive results of a deep convolutional network~\cite{krizhevsky2012imagenet} on the ILSVRC 2010 and 2012 datasets~\cite{russakovsky2015imagenet}, in the DL field a dramatic increase in the adaptation of neural networks to a broad set of visual recognition problems has been observed.
%These problems include image classification~\cite{khan2020survey}, image segmentation~\cite{minaee2021image}, object detection~\cite{liu2020deep} and other~\cite{ioannidou2017deep,litjens2017survey,shorten2019survey,zhu2020dark}.

The domains most related to AVR are presented in Fig.~\ref{fig:taxonomy}. All of them address the problem of reasoning about abstract concepts present in the image or identifying hidden rules (also referred to as patterns) that govern visual entities, although in various settings.
First of all, such tasks often emerge in the field of Visual Question Answering (VQA)~\cite{antol2015vqa,wu2017visual,kafle2017visual,manmadhan2020visual,srivastava2020visual}, where the goal is to answer a question written in natural language referring to an associated image.
In VQA, the information present in the image is sufficient for answering the related questions, whereas the Visual Commonsense Reasoning (VCR) field~\cite{yu2016modeling,kahou2017figureqa,suhr2019corpus,zellers2019recognition,xie2019visual} takes it a step further and places the tasks in real-world settings, where external knowledge is often required to solve them.
The concept of real-world settings is also utilised in the Physical Reasoning (PR) problems~\cite{santoro2017simple,bakhtin2019phyre,allen2020rapid,riochet2020intphys,baradel2020cophy} that comprise environments governed by underlying physical mechanisms or rules.

In contrast, AVR does not depend on real-world knowledge -- in fact, it even assumes that the test-taker will be unfamiliar with the presented shapes and their attributes.
Furthermore, AVR matrices are not associated with instance-specific questions, but rather a general short description of the whole problem is provided.
In comparison to PR problems where objects are governed by physics, AVR puzzles rely on logical relations or abstract rules defined through visual analogies.

A few existing works have already attempted to review the progress in constructing and solving AVR tasks using classical~\cite{hernandez2016computer} and more recent~\cite{mitchell2021abstraction,van2021much,stabinger2021evaluating} methods.
In~\cite{mitchell2021abstraction}, the author discusses recent progress in AI systems toward analogy-making and conceptual abstraction, and shows how far the current approaches are from the human level.
The work puts into question the role of the popular RPM benchmarks in measuring the analogy-making capacity due to their large training sets and advocates for relying on BPs and Abstraction and Reasoning Corpus (ARC)~\cite{chollet2019measure} that offer few-shot learning setups, which are more aligned with how analogy-making process is evaluated in humans.
A similar perspective is taken in~\cite{van2021much}, where the authors refer to recent works on RPMs and BPs as demonstrations of how limited current AI systems are --- they are able to solve single benchmarks at most and rarely generalise to other problems even from the same domain.
The topic of relational reasoning and concept learning is further explored in~\cite{stabinger2021evaluating}, where the shortcomings of current ML approaches to solving AVR tasks and limitations of selected benchmarks are discussed.

In contrast, in this work we refrain from focusing on exemplary benchmarks and instead provide a comprehensive review of available AVR tasks that were proposed to date, some of them not mentioned in the existing surveys.
We align the AVR problems with the proposed taxonomy which should be helpful in making informed decisions about which AVR benchmarks should be selected in future works.
While prior works mainly criticise AVR benchmarks due to their misalignment with approaches used for evaluating human analogy-making abilities, this review presents a more optimistic view of the field as a whole, describes fundamental challenges that are posed by AVR tasks in the context of computational intelligence, and advocates for their significance in other streams of research.

Our review also differs from similar prior works in the choice of described methods for solving AVR tasks.
While~\cite{mitchell2021abstraction} discusses progress across 3 fundamental approaches: symbolic methods, probabilistic program induction, and deep learning, the author reviews only a handful of proposed models.
Similarly, selected chosen methods are described in~\cite{stabinger2021evaluating} and their shortcomings are outlined.
The authors additionally hypothesize that attention mechanisms may be used as remediation.
Moreover,~\cite{van2021much} discusses general progress in deep learning and reinforcement learning with no specific focus on AVR domain.

Instead of advocating for a single approach or reviewing selected examples from a wide category of methods, in this survey we concentrate on the most promising recent DL models.
This allows to cover a wide spectrum of approaches and provides a unified view of recent strategies for designing DL-based AVR solvers.

\subsection{Contribution}
This survey describes both the leading AVR benchmarks on which the state-of-the-art was already advanced considerably, as well as emerging problems that highlighted the major shortcomings of current pattern analysis algorithms and are yet to be solved.
Overall, this paper:
\begin{itemize}
    \item comprehensively reviews emerging 
    %types of 
    AVR problems and benchmarks;
    \item introduces the AVR taxonomy that aligns existing AVR tasks in terms of their commonalities and differences;
    \item provides a unified view of the current DL approaches to solving AVR problems, which is grounded in the introduced taxonomy;
    \item links advances in AVR to other fields of practical application and presents possible future research paths.
\end{itemize}

\subsection{Structure}
We continue the survey by reviewing existing AVR problems and datasets in Section~\ref{sec:emerging-avr-problems}.
In Section~\ref{sec:avr-task-categorization} we categorise the outlined AVR tasks according to the proposed taxonomy.
Referring to the above categorisation, in Section~\ref{sec:overview-of-avr-models} we review a variety of approaches that were proposed to solve these tasks, with the main focus on DL methods.
Section~\ref{sec:discussion} presents a summary discussion of the AVR literature, identifies connections between AVR and other fields, and presents ideas for future work.
Section~\ref{sec:conclusion} concludes the survey.

    \begin{figure}[t]
    \centering
    \subfloat{\includegraphics[width=.45\textwidth]{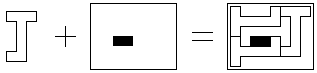}}
    \\
    \subfloat{\includegraphics[width=.45\textwidth]{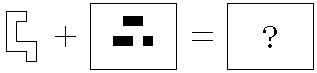}}
    \caption{\textbf{Octomino puzzle.}
    The goal of the puzzle is to arrange 4 copies of the octomino (a union of 8 unit squares) on the central image, such that each pair of shapes shares a common border.
    The black rectangles are unmovable and the shapes can't be placed on top of them.
    The problem appeared in~\cite{gardner2006colossal} and was adapted in the figure to be consistent with other discussed AVR tasks.
    While the rule that determines the arrangement of the octominos in the upper matrix might be discovered by humans after some consideration, current AVR benchmarks of this type are yet much simpler.
    Nevertheless, they pose a significant challenge to ML approaches.}
    \label{fig:human-avr-puzzle}
\end{figure}

\section{Emerging AVR problems}\label{sec:emerging-avr-problems}

\begin{figure*}[ht]
    \centering
    \subfloat[Sandia~\cite{matzen2010recreating}]{\includegraphics[width=.18\textwidth]{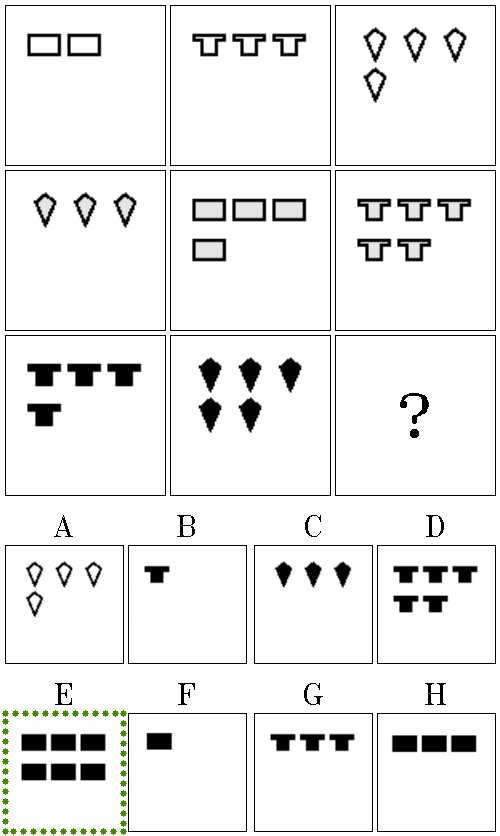}\label{fig:rpm-sandia}}
    ~
    \subfloat[Synthetic~\cite{wang2015automatic}]{\includegraphics[width=.18\textwidth]{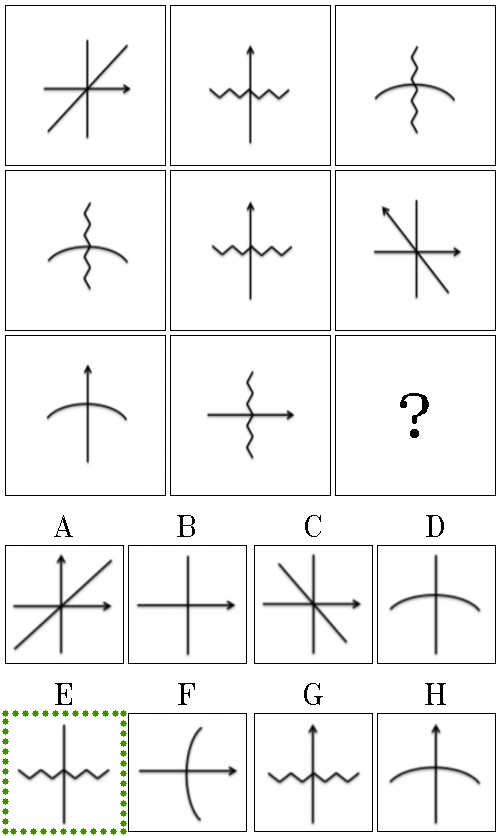}\label{fig:rpm-wang2015automatic}}
    ~
    \subfloat[G-set~\cite{mandziuk2019deepiq}]{\includegraphics[width=.18\textwidth]{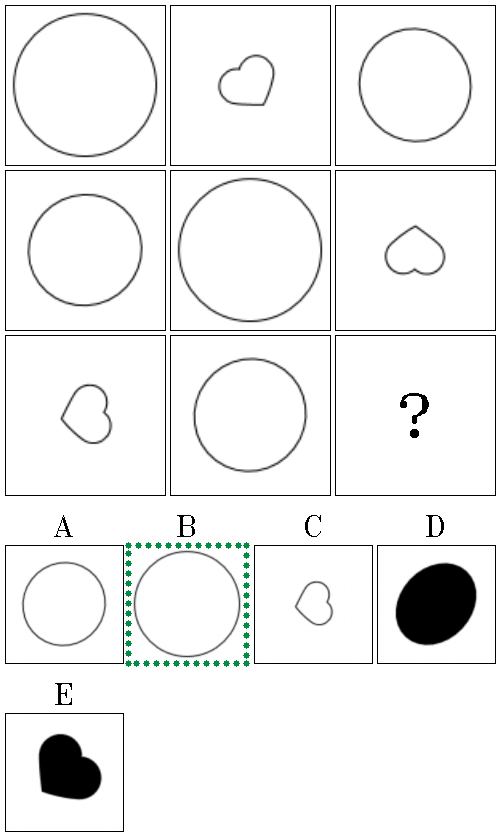}\label{fig:rpm-deepiq}}
    ~
    \subfloat[PGM~\cite{barrett2018measuring}]{\includegraphics[width=.18\textwidth]{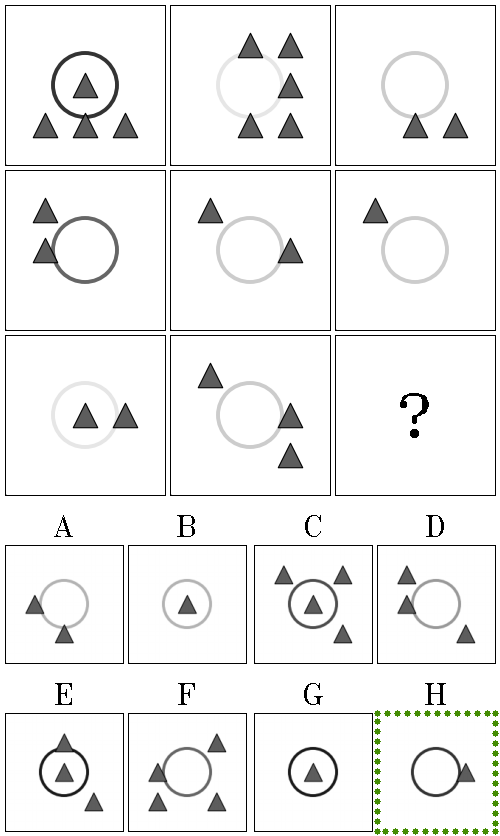}\label{fig:rpm-pgm}}
    ~
    \subfloat[I-RAVEN~\cite{zhang2019raven,hu2021stratified}]{\includegraphics[width=.18\textwidth]{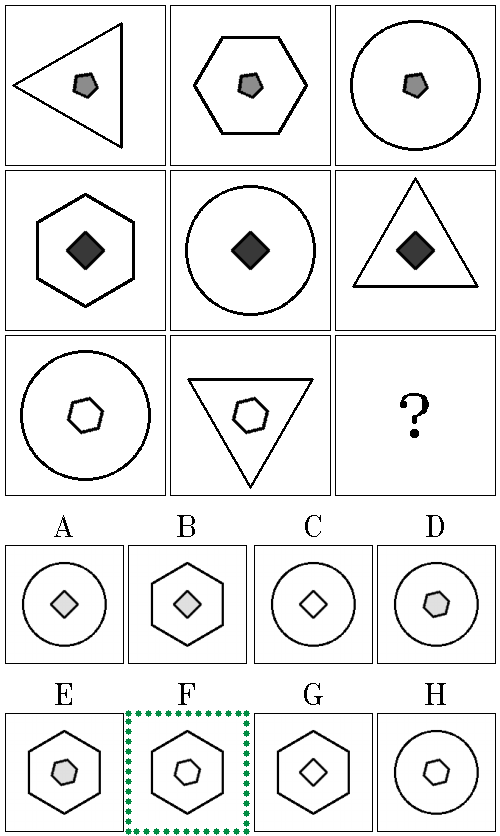}\label{fig:rpm-i-raven}}
    \caption{\textbf{Raven's Progressive Matrices.}
    To solve the task, the $3 \times 3$ grid of images has to be completed with one of the answer panels (A-E in (c) and A-H in the other cases).
    Matrices from RAVEN, I-RAVEN and RAVEN-FAIR differ only in the way of generating answers, hence only a selected matrix from the I-RAVEN dataset is shown.
    Correct answers are marked with a dotted green boundary.
    These answers can be identified as follows:
    (a) in each row, the number of objects increases by 1 from left to right;
    (b) in each row, any two panels differ in the shapes they are composed of (a given line/arrow/zigzag/arc appears in at most 1 image in a given row);
    (c) each row has 2 white circles and one is bigger than the other;
    (d) in each row, the position of the triangle in the right column can be obtained by applying logical AND to triangle positions in the preceding columns (circles in the background are so-called \emph{distractors} and their role is to make the task more difficult);
    (e) each row has three different outer shapes (a triangle, a hexagon, and a circle) and constant inner shape in the same color.}
    \label{fig:rpm-examples}
\end{figure*}

Even though there exists a wide landscape of AVR tasks and logic puzzles designed for human solvers~\cite{gardner2005martin,gardner2006colossal,dudeney2016536}, e.g. popular Octomino puzzles (Fig.~\ref{fig:human-avr-puzzle}), the majority of current DL research in this area is unquestionably devoted to solving RPMs~\cite{raven1936mental,raven1998raven,malkinski2022deep}.
Recent works, however, start to consider other than RPM formulations of AVR problems that allow to assess and analyse the capabilities of DL approaches from complementary perspectives.

In general, an AVR problem consists of a set of images that depict simple 2D shapes, which differ in some visual attributes.
The images are arranged in a meaningful problem-specific structure that is often realised in some form of a grid.
At the heart of these tasks lie the abstract rules that describe the relations between images, the encompassed objects, and their attributes.
When solving an AVR task, the main goal is to discover this underlying structure that can later be applied to provide the final answer.
The method of providing the answer depends on the problem and can be formulated in several setups such as classification, generation, or description in natural language.
In the remainder of this section, we describe contemporary AVR tasks one by one and then align them with the AVR taxonomy in Section~\ref{sec:avr-task-categorization}.

\subsection{Raven's Progressive Matrices}\label{sec:rpm}

Being the most often discussed AVR task in DL literature, RPMs were chosen as a proxy for studying machine intelligence for a plethora of automatic pattern analysis algorithms.
To a large extent, this comes from their wide-spread presence in the cognitive literature, where RPMs were found to be well-suited for testing human intelligence~\cite{carpenter1990one} and highly diagnostic of human abstract and relational reasoning abilities~\cite{snow1984topography}.

Another key contributor to the number of DL publications that tackle RPMs is a wide suite of their automatic generation methods.
Despite a relatively small---from DL perspective---number of instances in the initial set of hand-crafted RPMs~\cite{raven1936mental,raven1998raven}, automatic methods allow generating a huge number of instances that are required for current data-hungry ML methods.
Preliminary works include Sandia matrix generation software~\cite{matzen2010recreating}, synthetic RPMs~\cite{wang2015automatic}, and G-set -- a set of RPMs that was used to train the DeepIQ system~\cite{mandziuk2019deepiq}.
To better understand the shortcomings of current ML approaches in solving RPMs, PGM~\cite{barrett2018measuring} and RAVEN~\cite{zhang2019raven} datasets were introduced that enable measuring performance in specific generalisation regimes and contain images with compositional structure.
Apparently, the answer panels in RAVEN dataset were biased, which was discovered and mitigated in subsequent works that introduced I-RAVEN~\cite{hu2021stratified} and RAVEN-FAIR~\cite{benny2020scale}.
Example matrices from RPM datasets are shown in Fig.~\ref{fig:rpm-examples}.

Classical Raven's matrices consist of two components.
The set of context panels contains 8 images arranged in a $3 \times 3$ grid with a missing bottom-right image.
The goal of the test-taker is to choose a matching image from the second component -- a set of up to 8 answer panels.
To select the correct answer, it is required to identify hidden rules (e.g. constancy, progression, logical operators such as AND or XOR) that govern the shapes (e.g. circle, star, triangle) and their attributes (e.g. color, rotation, size) present in the images.
Depending on the dataset, the rules may be applied row-wise, column-wise, or---sometimes---also diagonally.
After identifying the rules and the way they are applied, an answer that preserves all the rules, after being placed in the bottom-right panel, has to be chosen.
In the default setup, realised in all the datasets described in this work, there is only a single answer panel that preserves all the rules, whereas the remaining ones satisfy fewer number of rules or none of them.

RPMs are the subject of a recent survey, where existing benchmarks, DL models and learning methods are reviewed~\cite{malkinski2022deep}.
The authors also discuss how RPMs facilitate development of successful methods that address practical applications.
We further explore this topic in Section~\ref{sec:discussion}, where we argue that not only RPMs, but also other AVR tasks are relevant and influential for various research streams.

\begin{figure*}[t]
    \centering
    \subfloat[OR from \emph{shape-position} to \emph{line-type}]{\includegraphics[width=.32\textwidth]{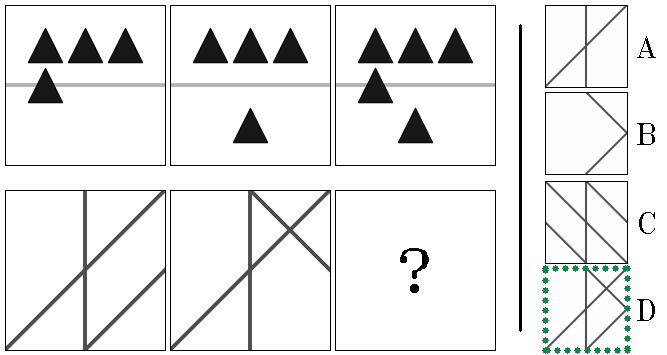}}
    \hfil
    \subfloat[AND from \emph{shape-position} to \emph{line-type}]{\includegraphics[width=.32\textwidth]{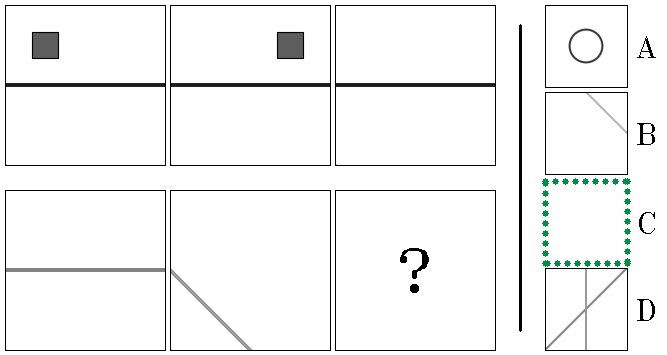}}
    \hfil
    \subfloat[XOR from \emph{shape-type} to \emph{line-colour}]{\includegraphics[width=.32\textwidth]{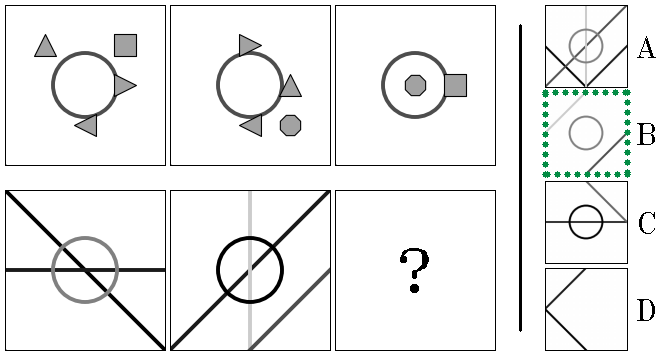}}
    \caption{
    \textbf{Visual analogy problems~\cite{hill2019learning}.}
    The $2 \times 3$ grid has to be completed with a matching answer panel selected from the right column of images labelled A-D.
    The relation expressed in the source domain (top row) has to be satisfied in the target domain (bottom row).
    Correct answers are marked with a dotted green boundary.
    In (a) and (b), logical rules OR and AND, respectively have to be identified in the \textit{shape-position} source domain and transferred to the \textit{line-type} target domain.
    Example (c) presents XOR applied to the source domain of \textit{shape-type}, which has to be transferred to \textit{line-colour} target domain (best viewed in electronic version of the paper).
    Note that some objects and attributes act as \emph{distractors} and their role is to make the task more difficult, e.g. horizontal lines (in (a) and (b)) or circles (in (c)) in the top rows.
    }
    \label{fig:visual-analogy-problems}
\end{figure*}

\subsection{Visual Analogy Problems}\label{sec:VAP}
In RPM tasks the solver has to identify abstract patterns that govern the objects and their attributes and select an answer for which the same rules are instantiated.
However, the images in an RPM are oftentimes similar -- one can say they come from the same domain.
On the other hand, the ability to make analogies is best verified when a given rule or concept from a source domain has to be applied to entities from a different target domain~\cite{gentner1983structure,hofstadter1995fluid}.
Recently, several benchmarks with Visual Analogy Problems (VAPs) that facilitate such a setting have been proposed.

In general, VAPs are composed of several panels divided into two parts.
The first part (source domain) presents a relation, which has to be reinstantiated in the second part (target domain), e.g. by completing or inpainting a missing panel.
Most common in the literature are VAPs composed of 4 panels, where the final completed matrix should satisfy the following symmetry: A is to B, as C is to D (also written as A:B::C:D).
However, emerging AVR benchmarks additionally consider extended problem formulations.

To verify the capabilities of DL models in such analogy making setting, a novel instantiaion of VAPs was proposed~\cite{hill2019learning}.
VAPs of this type are structurally similar to RPMs, i.e. they contain a set of context images arranged in a $2 \times 3$ grid with a missing bottom-right panel and the test-taker is supposed to complete the matrix with an appropriate answer panel chosen from a set of 4 candidate images.
Exemplary VAPs from~\cite{hill2019learning} are shown in Fig.~\ref{fig:visual-analogy-problems}.

The process of making analogies is often difficult, as it requires to encapsulate a concept implemented in a source domain in such an abstract way that would allow its flexible application to different target domains.
Remarkably, the authors of the dataset discovered that even simple DL models can be equipped with this ability given appropriate training setup.

To this end, they propose a method of learning analogies by contrasting abstract structure (LABC)~\cite{hill2019learning}. LABC generates the set of candidate panels in a deliberate way, that determines difficulty of the matrix.
The method differentiates between two kinds of candidates:
\emph{perceptually plausible} panels 
%-- that 
simply belong to the target domain;
\emph{semantically plausible} panels 
%-- that 
not only match the target domain, but in addition make the bottom row satisfy some relation after being completed with this panel.
However, only the correct answer makes the bottom row satisfy the same rule as demonstrated in the source domain.
This deliberate way of pairing the correct answer with semantically plausible candidates discourages shortcut solutions -- the underlying method is unable to choose the correct answer simply by checking if the candidate matches the target domain (perceptually plausible), or if the bottom row completed with the candidate satisfies any rule (semantically plausible), but also the rule has to match the one presented in the source domain.

The resultant learning process facilitates \textit{structural alignment} that was shown to be of crucial importance when transferring knowledge between problems~\cite{catrambone1989overcoming,gentner2001structural,bassok2003analogical}.
The VAPs dataset~\cite{hill2019learning} contains matrices with only perceptually plausible answers, as well as matrices with semantically plausible answers to test how the evaluated methods perform in a more demanding regime.

\begin{figure}[t]
    \centering
    \subfloat[Common features: light grey colour. B has darker colour.]{\includegraphics[width=.36\textwidth]{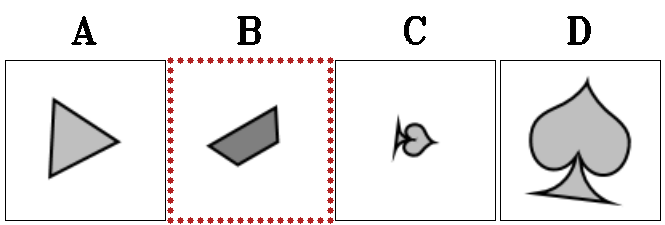}\label{fig:odd-one-out-1}}
    \\
    \subfloat[Common features: diamond shape. B is a triangle.]{\includegraphics[width=.45\textwidth]{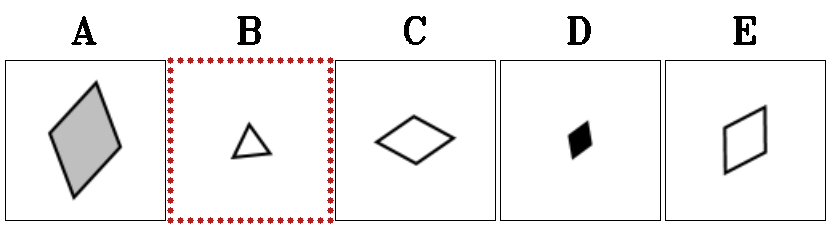}\label{fig:odd-one-out-2}}
    \\
    \subfloat[Common features: medium size. C is of small size.]{\includegraphics[width=.45\textwidth]{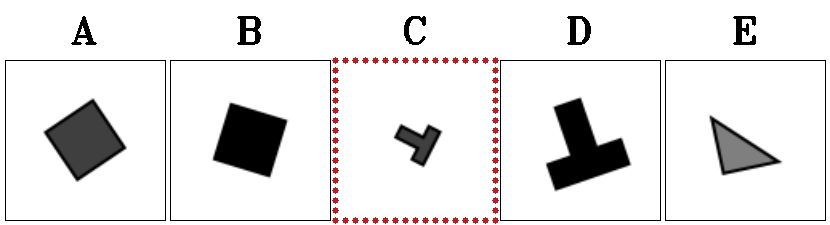}\label{fig:odd-one-out-3}}
    \\
    \caption{\textbf{Odd-one-out problems~\cite{mandziuk2019deepiq}.}
    Each problem instance contains a set of images, out of which one differentiates from the rest (see description in Section~\ref{sec:O-O-O}).
    Odd elements are marked with a dotted red boundary, with an explanation provided in the problem's caption.
    The matrices come from the G1-set~\cite{mandziuk2019deepiq}.}
    \label{fig:odd-one-out}
\end{figure}

\subsection{Odd-one-out}
\label{sec:O-O-O}
In both RPMs and VAPs the goal is to complete a set of images with a single matching panel that preserves the hidden rules that govern the objects and their properties.
In the odd-one-out ($\mathrm{O}^3$) problems, the set of images is already extended with a panel that breaks the pattern -- the odd one.
$\mathrm{O}^3$ problems have been widely studied from both cognitive and computational angles~\cite{dehaene2006core,gollin1972relational,lovett2008modeling,lovett2011cultural,ruiz2011building} and commonly appear in human puzzles~\cite{gardner2006colossal}.
In fact, the challenge of identifying an odd element in a set of objects has been recognised as fundamental not only to humans, but also to other animals~\cite{zentall1974comparison,zentall1980oddity}.

Despite crucial importance, $\mathrm{O}^3$ problems have been considered only recently as a benchmark for ML methods.
In~\cite{mandziuk2019deepiq}, the authors created two sets---G1-set and S1-set---of $\mathrm{O}^3$ matrices using images analogous to RPM datasets -- G-set (Fig.~\ref{fig:rpm-deepiq}) and S-set, respectively.
Examples of $\mathrm{O}^3$ tasks are shown in Fig.~\ref{fig:odd-one-out}.
Each problem instance is composed of $n \in \{4, 5, 6\}$ images that depict a single centrally-located object.
Each instance contains a subset of $n - 1$ panels, that share up to 3 common properties among $\{$color, rotation, shape, size$\}$ and there are no other $n-1$ panels that would share the same or greater number of features.
The odd element is defined as the one that differs in each of these shared properties (see Fig.~\ref{fig:odd-one-out}).

By employing similar objects as in G-set and S-set RPM benchmarks, the G1-set and S1-set datasets offer a unique testbed for evaluating transfer learning capacity of the tested methods.

\begin{figure*}[t]
    \centering
    \subfloat[Vertical vs horizontal]{\includegraphics[width=.32\textwidth]{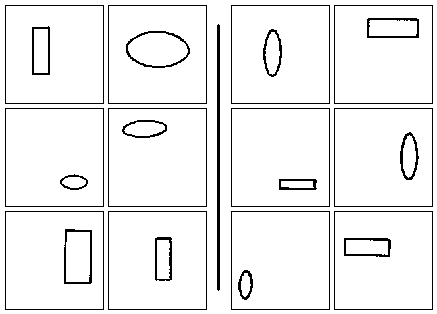}}
    \hfil
    \subfloat[Closed vs open]{\includegraphics[width=.32\textwidth]{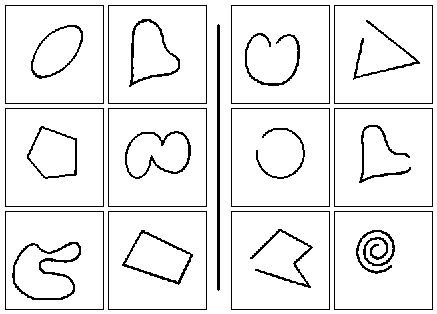}}
    \hfil
    \subfloat[Clockwise vs counterclockwise]{\includegraphics[width=.32\textwidth]{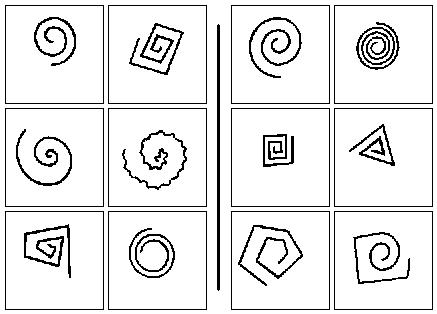}\label{fig:bongard-logo-matrix-1}}
    \\
    \subfloat[Same vs different shapes]{\includegraphics[width=.32\textwidth]{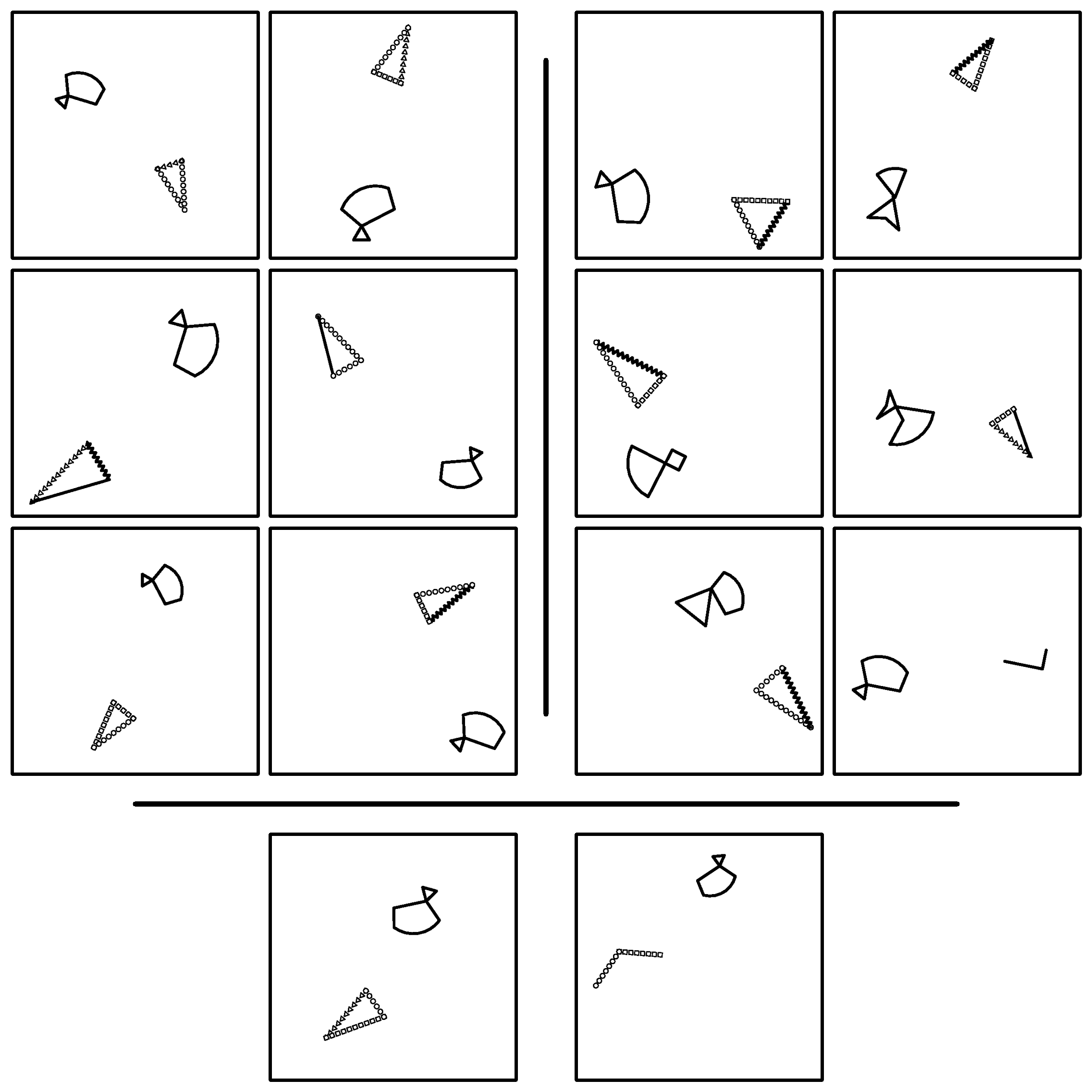}\label{fig:bongard-logo-matrix-2}}
    \hfil
    \subfloat[Circle and an arc of a semicircle]{\includegraphics[width=.32\textwidth]{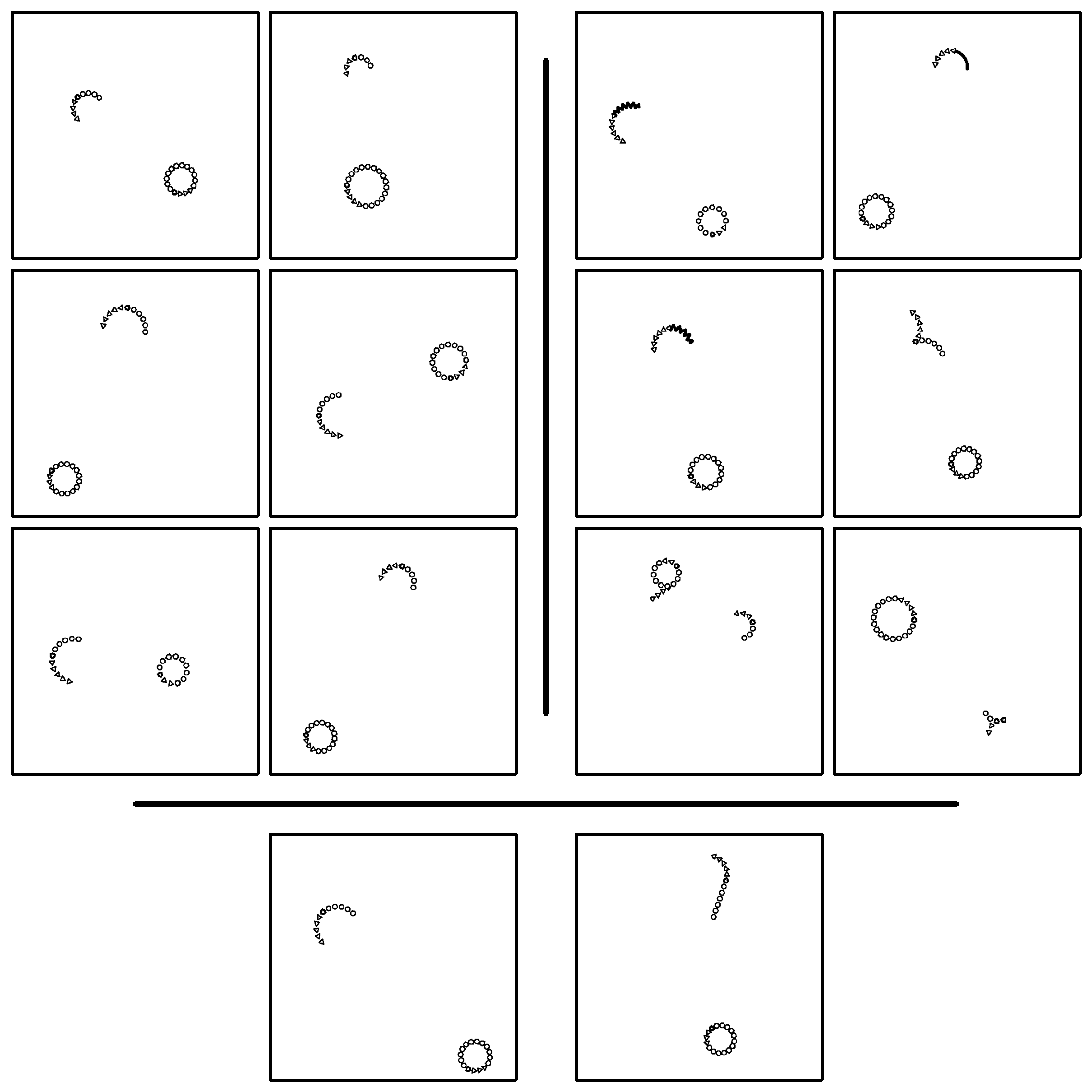}}
    \hfil
    \subfloat[Two convex shapes with a straight line]{\includegraphics[width=.32\textwidth]{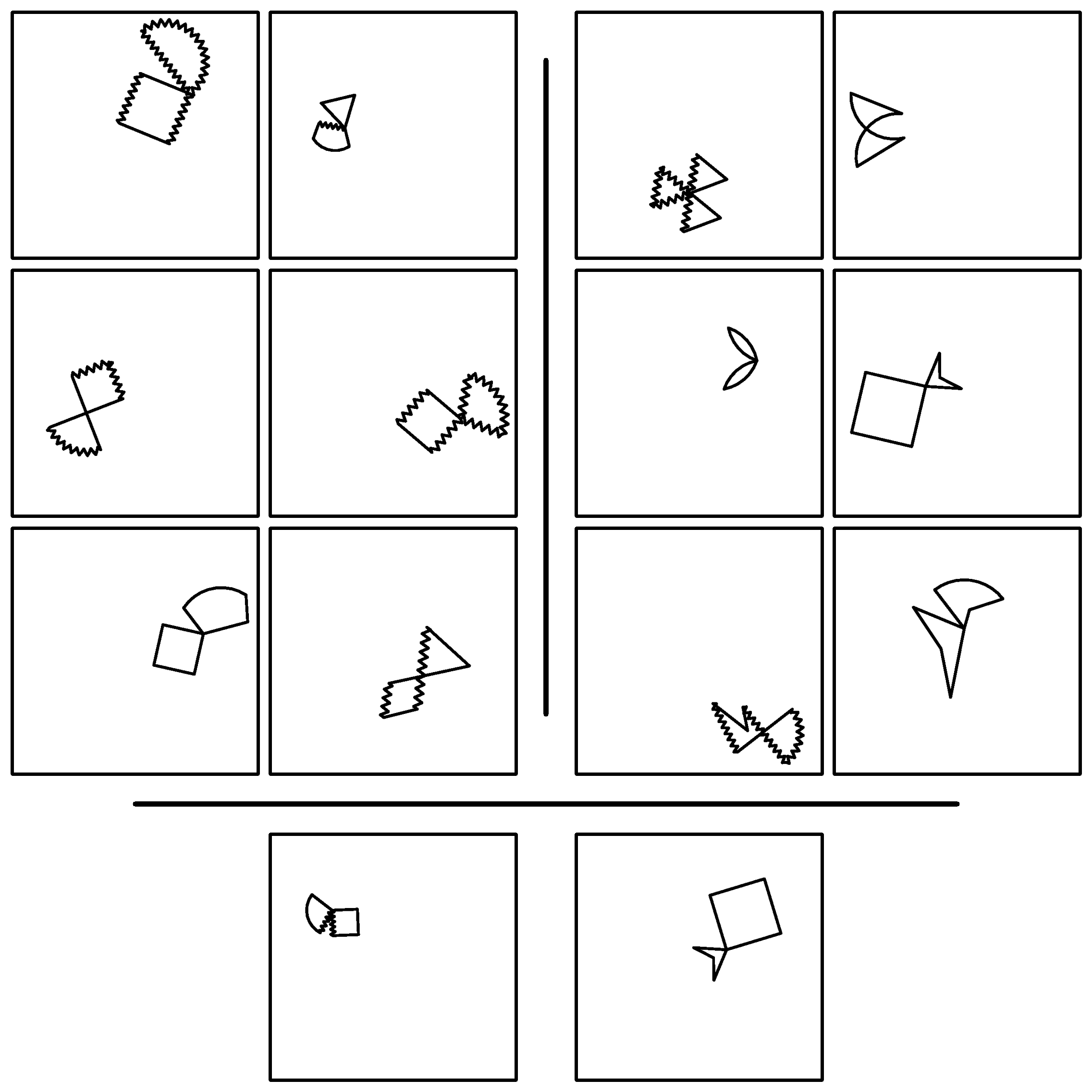}\label{fig:bongard-logo-matrix-3}}
    \caption{
    \textbf{Bongard problems~\cite{bongard1968recognition,nie2020bongard}.}
    The top row presents selected matrices from the hand-crafted set of BPs~\cite{bongard1968recognition}.
    To solve them, the abstract rules that separate left and right matrix parts have to be identified and described in natural language.
    The bottom row presents a simplified task that is more attainable for contemporary learning systems.
    Instead of describing the rules, two additional test images have to be assigned to either left or right matrix part, depending on which pattern they follow.
    The answers can be identified as follows:
    a) vertical rectangles and horizontal ellipses are on the left side, horizontal rectangles and vertical ellipses are on the right side;
    b) closed lines are on the left side, open lines are on the right side;
    c) clockwise spirals are on the left side, counterclockwise spirals are on the right side;
    d) each image on the left side presents two shapes: a separate acute triangle that is not equilateral, and a small equilateral triangle attached to a quadrangle with the longest side replaced by an arc, whereas each image on the right side contains only one of these two shapes;
    e) left part contains images with a circle and an arc of a semicircle (the arc is composed only of smaller circles and triangles), whereas right part contains images without the arc of a semicircle or with an arc composed of different elements (best viewed in electronic version of the paper under magnification);
    f) all shapes in images on the left side have two convex shapes with at least one straight edge and the right side contains images with either a concave shape, without straight edges, or with other than two shapes.}
    \label{fig:bongard-problems}
\end{figure*}

\subsection{Bongard Problems}\label{sec:bongard-problems}
Although some prior works have shown an impressive performance of DL models in solving RPMs~\cite{wu2020scattering,malkinski2022deep}, it was also identified that such models may struggle when the availability of data is scarce~\cite{zhuo2021effective}.
Moreover, recent literature reviews have criticized existing RPM benchmarks for their large sets of training instances, which casts doubt on the broader impact of the reported improvements in solving RPMs~\cite{mitchell2021abstraction,stabinger2021evaluating}.
On the other hand, Bongard Problems (BPs)~\cite{bongard1968recognition} are composed of abstract shapes and rules, which allow forming a rich testbed for evaluating learning methods in a low-sample regime.

The initial BP matrices~\cite{bongard1968recognition} were hand-crafted and therefore only few hundred problems were manually constructed by individual contributors~\cite{foundalis2021index}.
BP examples are shown in Fig.~\ref{fig:bongard-problems}.
Each problem instance is composed of two parts, each of them containing 6 images arranged in a grid.
Each part is governed by a distinct abstract rule.
The test-taker has to discover these rules that differentiate both parts.
Contrary to the already discussed problems, the rules in BPs are not sampled from a fixed, pre-defined set and additionally have to be described (by the test-taker/solving system) in natural language.

There have been two more approaches that framed the task of solving BPs to make it somehow attainable for the current learning systems.
In~\cite{kharagorgiev2018solving} the initial set of hand-crafted BPs~\cite{bongard1968recognition} is considered.
Each problem is framed as a binary classification task.
First, one image is extracted from each matrix part, and then for each of them, the model has to point out the part from which the image was extracted.
This contrasts with the problem formulation proposed in the seminal work~\cite{bongard1968recognition}, where matrices are solved by describing the separating rules in natural language.

A similar problem formulation is proposed in~\cite{nie2020bongard}, where the Bongard-LOGO benchmark is introduced.
The dataset measures the human-level concept learning and reasoning of AI agents with the help of $12\,000$ matrices.
Contrary to the approaches from~\cite{kharagorgiev2018solving,yun2020deeper}, for each problem, the test set is not extracted from the context matrices, but two additional test images are associated with each problem instance (see Figs.~\ref{fig:bongard-logo-matrix-1}--\ref{fig:bongard-logo-matrix-3}).
Moreover, Bongard-LOGO defines four testing splits that help to evaluate different generalization capabilities of the tested methods in a similar fashion to PGM regimes~\cite{barrett2018measuring} in the RPM task.
The abstract shapes presented in matrices from Bongard-LOGO are randomly generated by executing sequences of program instructions written in action-oriented LOGO language~\cite{abelson1974logo}.

Similarly to RAVEN, performance evaluation of human test-takers was conducted on Bongard-LOGO and compared to several meta-learning~\cite{ravi2016optimization,santoro2016meta} and DL methods~\cite{snell2017prototypical,mishra2018a,lee2019meta,raghu2020Rapid,chen2020new}.
The DL approaches demonstrated, to some extent, the ability to solve the BP matrices in a simpler problem setting, where the description of rules in natural language is replaced by a binary classification task.
However, a study on human solvers conducted in~\cite{nie2020bongard}, revealed that even current state-of-the-art meta-learning methods lag far behind the human concept-learning performance. 

\newcommand*{\SectorRadius}{1.5ex}
\newcommand*{\SectorLineWidth}{.6pt}
\newcommand*{\slice}[2]{%
  \begin{pgfpicture}
    \pgfpathmoveto{\pgforigin}%
    \pgfpathlineto{\pgfpointpolar{#1}{\SectorRadius}}%
    \pgfarc{#1}{#2}{\SectorRadius}%
    \pgfpathclose
    \pgfsetlinewidth{\SectorLineWidth}%
    \pgfusepath{stroke}%
  \end{pgfpicture}%
}

\begin{figure*}[t]
    \centering
    \subfloat[$? = 1$]{\includegraphics[width=.49\textwidth]{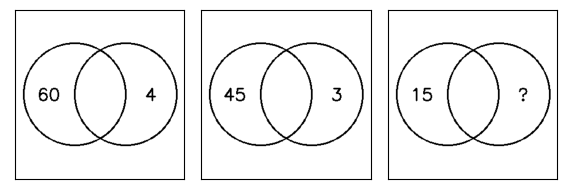}\label{fig:mns-1}}
    \hfil
    \subfloat[See the footnote for the answer.\textsuperscript{\tiny 2}]{\includegraphics[width=.49\textwidth]{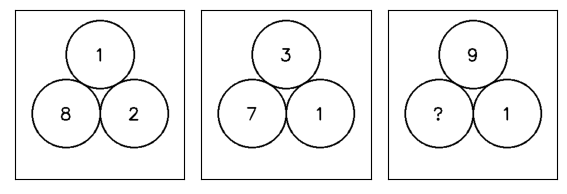}\label{fig:mns-2}}
    \\
    \subfloat[$? = 77$]{\includegraphics[width=.49\textwidth]{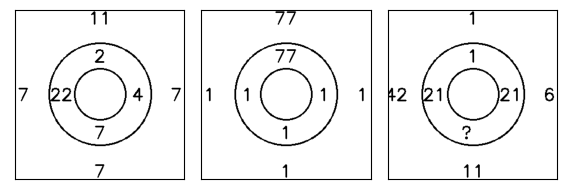}\label{fig:mns-3}}
    \hfil
    \subfloat[See the footnote for the answer.\textsuperscript{\tiny 2}]{\includegraphics[width=.49\textwidth]{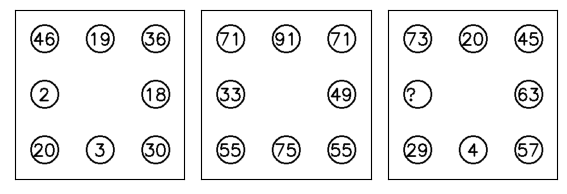}\label{fig:mns-4}}
    \\
    \subfloat[$? = 14$]{\includegraphics[width=.49\textwidth]{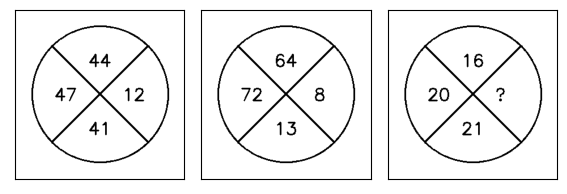}\label{fig:mns-5}}
    \hfil
    \subfloat[See the footnote for the answer.\textsuperscript{\tiny 2}]{\includegraphics[width=.49\textwidth]{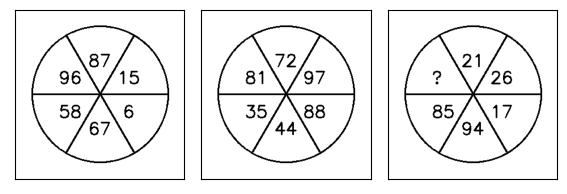}\label{fig:mns-6}}
    \caption{\textbf{Visual arithmetic problems~\cite{zhang2020machine}.}
    In order to solve these tasks, the test-taker has to identify an arithmetic expression presented in the first 2 panels and use it to find the value of a missing number in the rightmost panel.
    The answers can be deduced as follows:
    (a) in the left and center image, dividing left number by the right number gives 15, so $15 / ? = 15 \implies ? = 1$;
    (c) in both outer and inner ring, the following holds: $\mathrm{left} \times \mathrm{top} / \mathrm{right} \times \mathrm{bottom} = 77$, so $21 \times 1 / 21 \times ? = 77 \implies ? = 77$.
    (e) this involves performing the following arithmetic operation: $(\mathrm{left} - \mathrm{top}) \times \mathrm{right} + \mathrm{bottom}$, which for both left and center image gives 77, so $(20 - 16) \times ? + 21 = 77 \implies ? = 14$;
    To demonstrate the difficulty of these matrices, the answers to examples (b), (d) and (f) are hidden and the readers are encouraged to attempt solving them.
    The answers are provided in the footnote.\protect\footnotemark}
    \label{fig:machine-number-sense}
\end{figure*}

\subsection{Arithmetic Visual Reasoning}
The majority of abstract visual reasoning tasks require the solver to reason about abstract patterns composed of primitive objects (e.g. squares, triangles, stars, etc.) with certain attributes (size, rotation, location, fill-in pattern, etc.).
However, cognitive human studies have shown that there is another aspect of human intelligence that naturally emerges from vision, namely \textit{the sense of numbers}~\cite{dehaene2011number}, that refers to the understanding of numbers and associated operations, as well as the ability to solve the related mathematical (usually arithmetic) problems.
The sense of numbers was identified as a fundamental component of early human development~\cite{wynn1992addition,temple1998brain} and the interpretation of a magnitude expressed in a symbolic form was found to be highly indicative for human competency in solving diverse mathematical tasks~\cite{schneider2017associations}.

Although prior works have shown that neural models can be applied to mathematical reasoning problems expressed in the form of text~\cite{kushman2014learning,huang2016well,saxton2019analysing} or symbolic~\cite{lample2019deep} inputs, a recent work~\cite{zhang2020machine} has demonstrated substantial limitations of current DL approaches when simple arithmetic problems are integrated into AVR puzzles.
The authors introduced the Machine Number Sense (MNS) dataset~\cite{zhang2020machine} that evaluates the capability of intelligent systems to understand numerical symbols with relational operations between them (\textit{crystallised intelligence}), and in adaptive problem-solving (\textit{fluid intelligence}).

Each MNS problem is composed of 3 images of simple geometric figures.
The problems may come in 3 different types:
\emph{combination} integrates few geometric shapes into a specific spatial configuration -- in problems with 2 shapes, the structures overlap (Fig.~\ref{fig:mns-1}) or one is included in the other (Fig.~\ref{fig:mns-3}), while in problems with 3 shapes the structures are tangent to each other (Fig.~\ref{fig:mns-2});
\emph{composition} (Fig.~\ref{fig:mns-4}) arranges a set of small geometric shapes into a larger structure (a line, a cross, a triangle, a square, or a circle);
\emph{partition} (Figs.~\ref{fig:mns-5} and~\ref{fig:mns-6}) divides a single geometric shape into several parts.
The geometric shapes used in these problems are similar to the already discussed AVR tasks and include a triangle, a square, a circle, a hexagon, and a rectangle.
In contrast to other AVR problems, visual arithmetic matrices additionally contain integers from 1 to 99 located in certain pre-defined locations.
For a given problem instance, the test-taker is supposed to discover an arithmetic expression satisfied by the printed numbers and apply it to compute a missing number in the rightmost matrix panel.
The expression may use $4$ fundamental operators: addition, subtraction, multiplication, and division.

To measure the ability of DL models to solve such arithmetic visual problems, several baseline DL architectures were tested including CNN~\cite{lecun1990handwritten}, LSTM~\cite{hochreiter1997long}, RN~\cite{santoro2017simple}, and ResNet~\cite{he2016deep}.
All tested models struggled with solving the provided matrices and achieved results much worse than a group of human solvers -- on average, humans were able to solve 78\% of matrices from MNS, while the best performing DL model managed to tackle only 25\% of them.
Even though current DL approaches have demonstrated some ability to solve arithmetic tasks~\cite{kushman2014learning,huang2016well,saxton2019analysing,lample2019deep}, the above-described study~\cite{zhang2020machine} revealed that problems located at the intersection of mathematics and AVR remain generally challenging for ML/DL methods.
% \begin{figure}[t]
%     \centering
%     \includegraphics[width=.4\textwidth]{images/vaec/analogy_scale_test1_6000}
%     \caption{
%     \textbf{Visual analogy extrapolation problem~\cite{webb2020learning}.}
%     Given images $x_1$, $x_2$, $x_3$, the goal is to select the image $x_4$ from a set of up to 6 possible answers (in this example there are 4 choices, A-D) such that the relation between $x_3$ and $x_4$ is analogous in terms of brightness, size and location to the one between $x_1$ and $x_2$.
%     Correct answer is marked with a dotted green boundary.
%     The matrix is presented in greyscale for readability, but as described in~\cite{webb2020learning}, matrices from VAEC contain RGB images with green squares on a grey background.
%     }
%     \label{fig:vaec}
% \end{figure}

\begin{figure}[t]
    \centering
    \includegraphics[width=.4\textwidth]{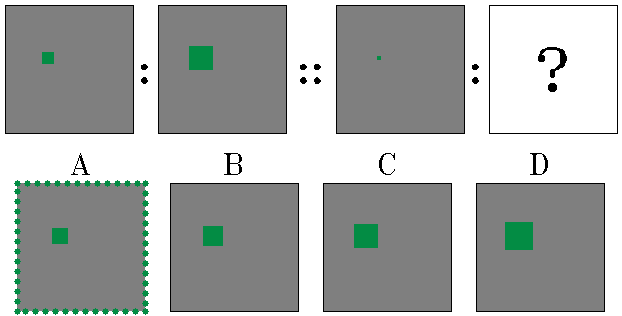}
    \caption{
    \textbf{Visual analogy extrapolation problem~\cite{webb2020learning}.}
    Given images $x_1$, $x_2$, $x_3$, the goal is to select the image $x_4$ from a set of up to 6 possible answers (in this example there are 4 choices, A-D) such that the relation between $x_3$ and $x_4$ is analogous in terms of brightness, size and location to the one between $x_1$ and $x_2$ (best viewed in electronic version of the paper under magnification).
    Correct answer is marked with a dotted green boundary.
    }
    \label{fig:vaec}
\end{figure}

\subsection{Visual Extrapolation Problems}
One of the reasons why current DL approaches still lag far behind humans in various AVR problems 
%partly arises from 
is their weak ability to extrapolate, which some works have argued is a key characteristic of human intelligence~\cite{lake2017building}.
The outcomes that support this thesis were emphasised in the context of solving RPMs, where DL models performed poorly in the \textit{extrapolation regime} of PGM dataset~\cite{barrett2018measuring,malkinski2020multilabel}.
Similar observations were made in VAPs from~\cite{hill2019learning} where, in general, the DL methods achieve weaker performance when extrapolating concepts to novel domains.

\begin{figure}[t]
    \centering
    \subfloat[An example of the train sequence. Square's area gradually decreases from 49 to 25 pixels. Its top-left corner shifts by 1 pixel right.]{\includegraphics[width=.45\textwidth]{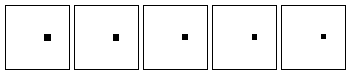}}
    \\
    \subfloat[Test sequence example. The area gradually increases from 121 to 729 pixels. Its top-left corner shifts by 5 pixels up and by 6 pixels left.]{\includegraphics[width=.45\textwidth]{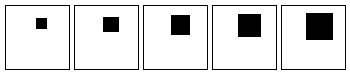}}
    \caption{
    \textbf{Dynamic object prediction task~\cite{webb2020learning}.}
    Given a sequence of $T - 1$ images of squares, the goal is to recreate the $T$'th (rightmost) image that follows the pattern.
    To test extrapolation capabilities, the training images contain only small squares with width $w \in [3, 13]$ (measured in pixels), whereas the test images contain large squares with $w \in [13, 31]$.
    The panels are of size $64 \times 64$ pixels.
    }
    \label{fig:dopt}
\end{figure}

To measure the extrapolation capabilities of DL methods,~\citet{webb2020learning} proposed two benchmark sets: the Visual Analogy Extrapolation Challenge (VAEC) and the dynamic object prediction task (DOPT) illustrated in Figs.~\ref{fig:vaec} and~\ref{fig:dopt}, respectively.
Both datasets contain problems that test various extrapolation realizations in terms of scale and translation.

% answers to MNS examples (b), (d), (f)
\footnotetext{Denoting L=left, R=right, T=top, B=bottom, C=center, the remaining answers are:\\
(b) L $+$ T $+$ R $= 11 \implies ? = 1$\\
(d) TL $-$ BR $=$ TR $-$ BL $=$ TC $-$ BC $=$ RC $-$ LC $= 16 \implies ? = 47$\\
(f) $\slice{180}{120} - \slice{120}{60} = \slice{60}{0} - \slice{300}{360} = \slice{300}{240} - \slice{240}{180} = 9 \implies ? = 30$}

Specifically, matrices from  VAEC present yet another instatiation of VAPs, which are oriented towards extrapolation -- given images $x_1$, $x_2$, $x_3$, a missing image $x_4$ has to be selected from a set of answers, such that the relation between $x_3$ and $x_4$ is analogous to the relation between $x_1$ and $x_2$.
Given 4 possible axis of variation (brightness, size, x coordinate, y coordinate), the relation between $x_1$ and $x_2$, as well as between $x_3$ and $x_4$ varies in the same direction and with the same magnitude.
The set of answers contains only such choices that vary along the same axis, however, only one image correctly completes the analogy.
This approach of defining available answer panels resembles the LABC method~\cite{hill2019learning} where the set of answers contains only semantically plausible candidates.
However, in contrast to VAPs from~\cite{hill2019learning}, matrices from VAEC explicitly focus on extrapolation in terms of brightness, scale and translation, while the benchmark proposed by~\citet{hill2019learning} focuses on forming analogies between different domains.

VAEC matrices fall into $2$ regimes that test different types of extrapolation: invariance to \emph{translation} in terms of size, brightness and location; and invariance to \emph{scale} between visual objects.
Problems in each regime are divided beforehand into train and test splits, which allows measuring generalisation of the tested methods to values from a much bigger range than encountered during the learning phase.

In addition, the authors proposed DOPT, a complementary extrapolation task.
Given a sequence of images $x_1$, \ldots, $x_{T-1}$ that depict a gradually changing and moving square observed at $T-1$ consecutive time steps, the task is to predict square's final size and position at time $T$ (see Fig.~\ref{fig:dopt}).
To form an extrapolation benchmark, train matrices depict only small squares with width $w \in [3, 13]$ (measured in pixels), while test matrices contain larger squares with $w \in [13, 31]$.

The authors of~\cite{webb2020learning} have shown that traditional DL approaches fail to demonstrate consistent performance in any of the two regimes.

To facilitate the extrapolation ability of neural models, the Temporal Context Normalization (TCN)~\cite{webb2020learning} method was proposed, which relies on a task-relevant temporal window over which the representations are normalised.
This approach allows neglecting absolute feature magnitudes while preserving the in-between feature relations.
The proposed normalisation technique excelled in both considered extrapolation tasks and additionally improved the state-of-the-art results on the VAP benchmark~\cite{hill2019learning} discussed in section~\ref{sec:VAP}.

\begin{figure}[t]
    \centering
    \subfloat[Problem type \#4]{\includegraphics[width=.22\textwidth]{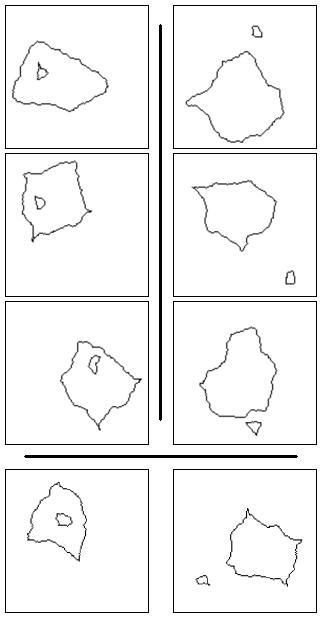}}
    ~
    ~
    \subfloat[Problem type \#21]{\includegraphics[width=.22\textwidth]{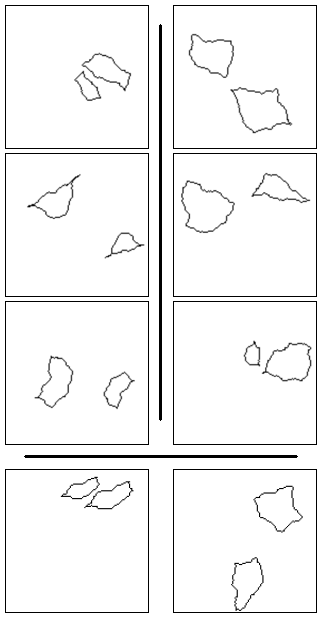}}
    \caption{
    \textbf{SVRT problems~\cite{fleuret2011comparing}.}
    To solve them, an abstract rule has to be discovered that differentiates images on the left side from those on the right side.
    Based on this rule, each test image from the bottom row has to be assigned to one of the image sides.
    In problem type \#4 (simple even for shallow CNNs) the smaller shape is inside the bigger one on the left side, while being outside on the right side.
    In problem type \#21 (demanding for simple CNNs) one of the shapes on the left side can be obtained by rotating, scaling and translating the other one, which is not the case for images on the right side.
    }
    \label{fig:svrt}
\end{figure}

\subsection{Same-different tasks}
In many real-world scenarios, one is often faced with a question whether a given object is \emph{same} or \emph{different} from another object.
This task is so fundamental for living beings that such discriminatory ability emerged in a variety of animal species~\cite{pepperberg1987acquisition,oden1990infant,wright2006mechanisms}.
The difficulty of this task rises when the differences are subtle, or when the concept of sameness concerns abstract properties of the compared objects.

In same-different (S-D) tasks from AVR domain, objects that are \emph{same} shouldn't be interpreted based on pixel-to-pixel similarity, but rather based on abstract visual concepts they present.
Consequently, \emph{different} objects not only differ in terms of individual pixels, but additionally present distinct---sometimes even opposite---abstract visual concepts.
Visual reasoning models for solving the S-D tasks received well-deserved attention in recent works~\cite{ricci2021same,forbus2021same}.

A seminal work~\cite{fleuret2011comparing} has shown that differentiating between same and different concepts has proven difficult to learn by the current ML models.
Namely,~\citet{fleuret2011comparing} proposed the Synthetic Visual Reasoning Test (SVRT) which comprises 23 types of classification problems.
Each problem type defines an abstract rule that differentiates images from 2 categories (left and right matrix parts).
The abstract rules are built on concepts such as symmetry, relative position, proximity, and a few more, with examples presented in Fig.~\ref{fig:svrt}.

\begin{figure*}[t]
    \centering
    \begin{tikzpicture}
        \begin{scope}[node distance=1.5cm,on grid]
            \node (p111) {\includegraphics[width=.14\textwidth]{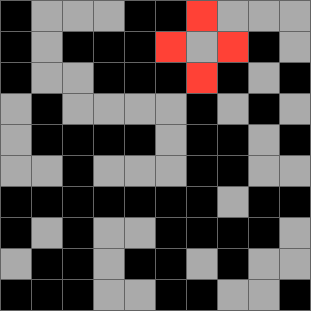}};
            \node (a11) [right=of p111] {\bigrightArrow};
            \node (p112) [right=of a11] {\includegraphics[width=.14\textwidth]{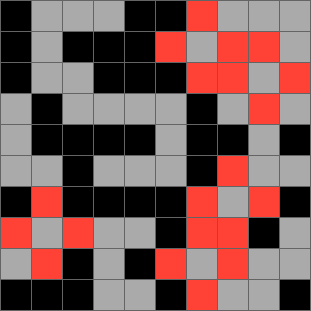}};
            
            \node (p121) [right=of p112, xshift=1.7cm] {\includegraphics[width=.14\textwidth]{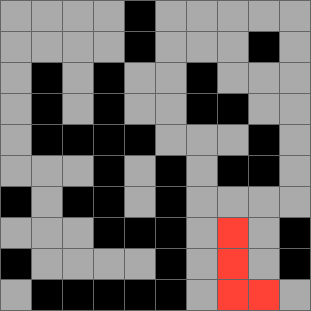}};
            \node (a12) [right=of p121] {\bigrightArrow};
            \node (p122) [right=of a12] {\includegraphics[width=.14\textwidth]{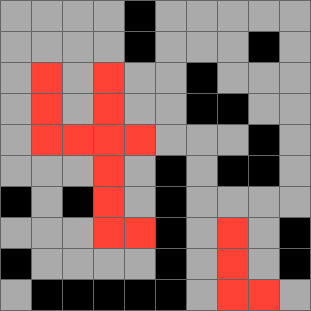}};
            
            \node (p131) [right=of p122, xshift=1.7cm] {\includegraphics[width=.14\textwidth]{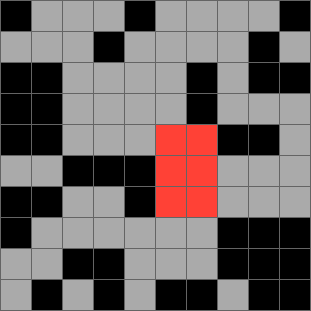}};
            \node (a13) [right=of p131] {\bigrightArrow};
            \node (p132) [right=of a13] {\includegraphics[width=.14\textwidth]{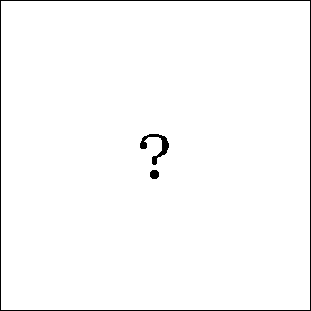}};

            \node (p211) [below=of p111, yshift=-1.4cm] {\includegraphics[width=.14\textwidth]{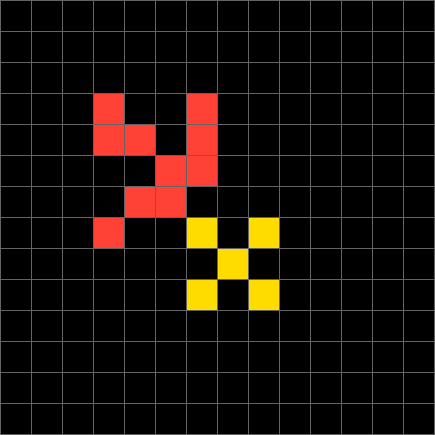}};
            \node (a21) [right=of p211] {\bigrightArrow};
            \node (p212) [right=of a21] {\includegraphics[width=.14\textwidth]{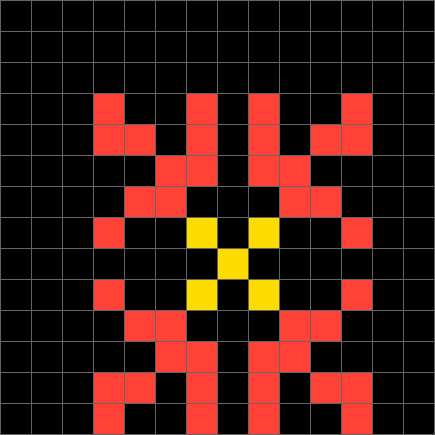}};
            
            \node (p221) [right=of p212, xshift=1.7cm] {\includegraphics[width=.14\textwidth]{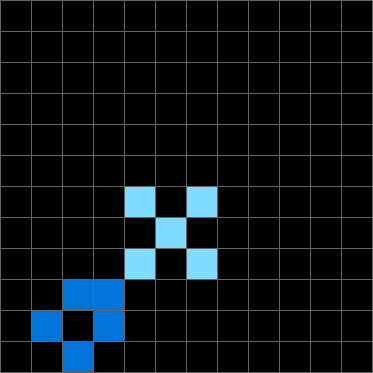}};
            \node (a22) [right=of p221] {\bigrightArrow};
            \node (p222) [right=of a22] {\includegraphics[width=.14\textwidth]{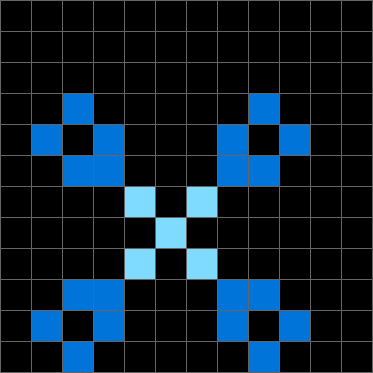}};
            
            \node (p231) [right=of p222, xshift=1.7cm] {\includegraphics[width=.14\textwidth]{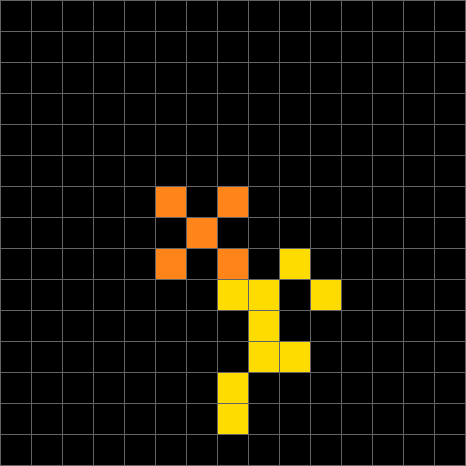}};
            \node (a23) [right=of p231] {\bigrightArrow};
            \node (p232) [right=of a23] {\includegraphics[width=.14\textwidth]{images/arc/missing}};
        \end{scope}
    \end{tikzpicture}
    \caption{
    \textbf{Abstraction and Reasoning Corpus problems~\cite{chollet2019measure}.}
    To solve tasks from ARC dataset~\cite{chollet2019measure}, the test-taker has to understand an abstract rule demonstrated by several examples and apply it to a new setting, by generating an output grid from scratch.
    In the problem presented in the top row, the red shape has to be copied into each available space (empty space is represented as black squares).
    In the bottom row, the goal is to extend the pattern, so that it will be symmetrical with respect to the central cross composed of 5 squares.
    }
    \label{fig:arc}
\end{figure*}

Given a problem type, a matrix is created by sampling images from both categories.
The number of sampled images determines the matrix difficulty -- the more images the matrix contains, the easier the abstract rule should be to discover.
Next, given a test image, initially not assigned to any of the 2 categories, the test-taker has to select the category to which the test image belongs.
The images present irregular randomly-generated black and white closed contours with no additional distracting attributes, which allows unbiased evaluation of human (or machine) reasoning skills irrespective of previous experience.

Preliminary attempts to tackle the SVRT challenge with DL systems demonstrated that traditional CNN architectures struggle with solving the S-D tasks~\cite{ellis2015unsupervised,stabinger201625}.
However, a parallel work on a related problem revealed that such unsatisfactory results may arise not from the weak expressive power of DL models, but rather from optimization difficulties and naive learning curriculum~\cite{gulccehre2016knowledge}.
By injecting task-relevant information into intermediate-level representations,~\citet{gulccehre2016knowledge} demonstrated that a seemingly weak model may achieve close to perfect results on these tasks, given appropriate supervision.

Motivated by these findings, subsequent works further explored the topic.
In~\cite{ricci2018same,kim2018not}, the authors analysed in detail the successes and failures of CNN architectures on the SVRT dataset.
It was discovered that the models handled categorization problems concerning the relative spatial position of shapes relatively well, while at the same time struggled with problems that involve rotation, translation, reflection, and identification of symmetries.
To facilitate a more fine-grained analysis of the problem, the authors introduced the parametric SVRT dataset (PSVRT~\cite{ricci2018same,kim2018not}), which remains an open challenge for DL systems until today.

On the other hand, even though it was concluded in~\cite{ricci2018same} that DL models struggle with solving S-D tasks, it was later shown in~\cite{messina2019testing,messina2021solving,funke2021five} that the initial categorization problems from SVRT can be solved with ResNets -- modern DL models that excel in computer vision.

Alternatively, in~\cite{bohn2019few} the authors show how a specific image pre-processing technique can make even CNN-based classifiers competitive in solving matrices from SVRT.

\subsection{Abstraction and Reasoning Corpus}
Though the described AVR benchmarks pose a real challenge for contemporary learning systems on their own, there is a fundamental difference in how these tests are used in measuring machine intelligence versus how they are used in psychometric tests for humans.
Namely, current benchmarks consist of huge training sets, that not only allow the trained systems to prepare in advance for the tasks they will encounter during the evaluation phase, but more crucially are known beforehand by the developers of these models.
As a result, current learning systems are hand-crafted for a particular task they will be evaluated on -- a topic we further explore in Sections~\ref{sec:overview-of-avr-models} and~\ref{sec:discussion}.
This critical difference between human and machine evaluation settings sparked a series of discussions in the AVR community~\cite{mitchell2021abstraction} and motivated the development of the Abstraction and Reasoning Corpus (ARC), with examples shown in Fig.~\ref{fig:arc}.

The dataset shifts the evaluation setting from test problems that are known in advance, into an uncharted territory more aligned with human psychometric tests where the types of tasks found in the training and test sets do not intersect.

More specifically, ARC comprises 1000 distinct AVR tasks divided into training (400 instances) and test (600 instances) sets.
Each task presents a unique challenge illustrated with several examples.
In each task, the goal is to understand the problem we are faced with, based on the analysis of the exemplary demonstrations, so as a task-specific test example can be solved.
While the method of solving the task is unique in each case, the tasks share a similar structure -- each instance contains an input and an output grid, where each cell is coloured with one from a fixed set of 10 possibilities.
Grids can be of variable size ranging from $1 \times 1$ up to $30 \times 30$, and the test output grid with an answer has to be created from scratch by the test-taker.

Compared to the previously described AVR tasks, ARC doesn't divide the problem instances into train and test sets based on specific object attributes, but rather by the task being solved.
Each task encountered in the testing phase is novel for the solver, which recreates a setting found in human psychometric tests and limits the amount of task-specific experience -- the specific skills required for solving the presented task can only be gathered from few demonstrations.

Crucially, tasks from ARC do not focus on measuring how well can the tested method learn particular skills, but rather measure how efficient the learning system is in acquiring new skills from merely a few examples.

    \begin{table*}[t]
    \centering
    \caption{
    \textbf{Alignment of AVR tasks with AVR taxonomy.}
    AVR problems and their corresponding benchmarks are catalogued under the following 5 dimensions of the AVR taxonomy: Input shapes, Hidden rules, Target task, Cognitive function, Specific challenge.
    }
    \label{tab:categorisation-of-avr-tasks}
    \begin{tabular}{ll|cc|cc|ccc|cc|ccc}
        \toprule
        \rowcolor{lightergray} \textbf{Problem} & \textbf{Dataset} & \rotatebox{90}{\textbf{Geometric shapes}} & \rotatebox{90}{\textbf{Abstract shapes}} & \rotatebox{90}{\textbf{Explicit rules}} & \rotatebox{90}{\textbf{Abstract rules}} & \rotatebox{90}{\textbf{Classification}} & \rotatebox{90}{\textbf{Generation}} & \rotatebox{90}{\textbf{Description}} & \rotatebox{90}{\textbf{Completion}} & \rotatebox{90}{\textbf{Discrimination}} & \rotatebox{90}{\textbf{Domain transfer}} & \rotatebox{90}{\textbf{Extrapolation}} & \rotatebox{90}{\textbf{Arithmetic}} \\
        \midrule
        \multirow{5}{*}{\shortstack[l]{Raven's Progressive\\Matrices}} & Sandia~\cite{matzen2010recreating} & \checkmark & & \checkmark & & \checkmark &  & & \checkmark & \checkmark & & & \\
        & Synthetic~\cite{wang2015automatic} & \checkmark & & \checkmark & & \checkmark &  & & \checkmark & \checkmark & & & \\
        & G-set~\cite{mandziuk2019deepiq} & \checkmark & & \checkmark & & \checkmark &  & & \checkmark & \checkmark & & & \\
        & RAVEN~\cite{zhang2019raven,hu2021stratified,benny2020scale} & \checkmark & & \checkmark & & \checkmark &  & & \checkmark & \checkmark & \checkmark & & \\
        & PGM~\cite{barrett2018measuring} & \checkmark & & \checkmark & & \checkmark &  & & \checkmark & \checkmark & \checkmark & \checkmark & \\
        \midrule
        VAPs & \citet{hill2019learning} & \checkmark & & \checkmark & & \checkmark & & & \checkmark & \checkmark & \checkmark & \checkmark & \\
        \midrule
        \multirow{2}{*}{Bongard problems} & Hand-crafted~\cite{foundalis2006phaeaco} & & \checkmark & & \checkmark & & & \checkmark & & \checkmark & & & \\
        & Bongard-LOGO~\cite{nie2020bongard} & & \checkmark & & \checkmark & \checkmark & & & & \checkmark & & \checkmark & \\
        \midrule
        Same-different & SVRT~\cite{fleuret2011comparing} & & \checkmark & & \checkmark & \checkmark & & & & \checkmark & & & \\
        \midrule
        \multirow{2}{*}{Odd-one-out} & G1-set~\cite{mandziuk2019deepiq} & \checkmark & & \checkmark & & \checkmark & & & & \checkmark & \checkmark & & \\
        & S1-set~\cite{mandziuk2019deepiq} & \checkmark & & \checkmark & & \checkmark & & & & \checkmark & \checkmark & & \\
        \midrule
        Visual arithmetic reasoning & MNS~\cite{zhang2020machine} & \checkmark & & \checkmark & & \checkmark & & & \checkmark & & & & \checkmark \\
        \midrule
        \multirow{2}{*}{Extrapolation} & VAEC~\cite{webb2020learning} & \checkmark & & \checkmark & & \checkmark & & & \checkmark & \checkmark & & \checkmark & \\
        & DOPT~\cite{webb2020learning} & \checkmark & & \checkmark & & & \checkmark & & \checkmark & & & \checkmark & \\
        \midrule
        Abstraction and reasoning & ARC~\cite{chollet2019measure} & \checkmark & & & \checkmark & & \checkmark & & \checkmark & & \checkmark & \checkmark & \\
        \bottomrule
    \end{tabular}
\end{table*}

\begin{table}[t]
    \centering
    \caption{
    \textbf{Public datasets.}
    Majority of the discussed datasets are publicly available and can be found at the addresses listed below.
    Datasets marked with \faicon{github} are available on GitHub and can be accessed at the following address: https://github.com/\{address\}.
    }
    \label{tab:public-datasets}
    \small
    \begin{tabular}{>{\footnotesize}l|>{\footnotesize}l}
        \toprule
        \rowcolor{lightergray} \textbf{Dataset} & \textbf{Address} \\
        \midrule
        Sandia~\cite{matzen2010recreating} & \faicon{github} \href{https://github.com/LauraMatzen/Matrices}{LauraMatzen/Matrices} \\
        G-set~\cite{mandziuk2019deepiq} & \faicon{github} \href{https://github.com/deepiq/deepiq}{deepiq/deepiq} \\
        RAVEN~\cite{zhang2019raven} & \faicon{github} \href{https://github.com/WellyZhang/RAVEN}{WellyZhang/RAVEN} \\
        I-RAVEN~\cite{hu2021stratified} & \faicon{github} \href{https://github.com/husheng12345/SRAN}{husheng12345/SRAN} \\
        RAVEN-FAIR~\cite{benny2020scale} & \faicon{github} \href{https://github.com/yanivbenny/RAVEN\_FAIR}{yanivbenny/RAVEN\_FAIR} \\
        PGM~\cite{barrett2018measuring} & \faicon{github} \href{https://github.com/deepmind/abstract-reasoning-matrices}{deepmind/abstract-reasoning-matrices} \\
        \midrule
        \citet{hill2019learning} & \faicon{github} \href{https://github.com/deepmind/abstract-reasoning-matrices}{deepmind/abstract-reasoning-matrices} \\
        \midrule
        Hand-crafted~\cite{foundalis2006phaeaco} & \href{https://www.foundalis.com/res/bps/bpidx.htm}{foundalis.com/res/bps/bpidx.htm} \\
        Bongard-LOGO~\cite{nie2020bongard} & \faicon{github} \href{https://github.com/NVlabs/Bongard-LOGO}{NVlabs/Bongard-LOGO} \\
        \midrule
        SVRT~\cite{fleuret2011comparing} & \href{https://fleuret.org/git/svrt}{fleuret.org/git/svrt} \\
        \midrule
        G1-set~\cite{mandziuk2019deepiq} & \multirow{2}{*}{\faicon{github} \href{https://github.com/deepiq/deepiq}{deepiq/deepiq}} \\
        S1-set~\cite{mandziuk2019deepiq} & \\
        \midrule
        \multirow{2}{*}{MNS~\cite{zhang2020machine}} & \faicon{github} \href{https://github.com/zwh1999anne/Machine-Number-Sense-Dataset}{zwh1999anne/Machine-} \\
        & \href{https://github.com/zwh1999anne/Machine-Number-Sense-Dataset}{Number-Sense-Dataset} \\
        \midrule
        VAEC~\cite{webb2020learning} & \faicon{github} \href{https://github.com/taylorwwebb/learning\_representations\_that\_support\_extrapolation}{taylorwwebb/learning\_represen} \\
        DOPT~\cite{webb2020learning} & \href{https://github.com/taylorwwebb/learning\_representations\_that\_support\_extrapolation}{tations\_that\_support\_extrapolation} \\
        \midrule
        ARC~\cite{chollet2019measure} & \faicon{github} \href{https://github.com/fchollet/ARC}{fchollet/ARC} \\
        \bottomrule
    \end{tabular}
\end{table}

\section{Alignment of AVR tasks with the AVR taxonomy}\label{sec:avr-task-categorization}

To better understand the commonalities and differences between emerging AVR research tasks and the respective benchmarks, we further align them along the 5 dimensions of the introduced AVR taxonomy.
An aggregated summary is presented in Table~\ref{tab:categorisation-of-avr-tasks} and each dimension is discussed in detail in the following sections.
Table~\ref{tab:public-datasets} presents references to the publicly available datasets.

\subsection{Input shapes}
\begin{figure}[t]
    \centering
    \subfloat[Geometric shapes]{\includegraphics[width=.22\textwidth]{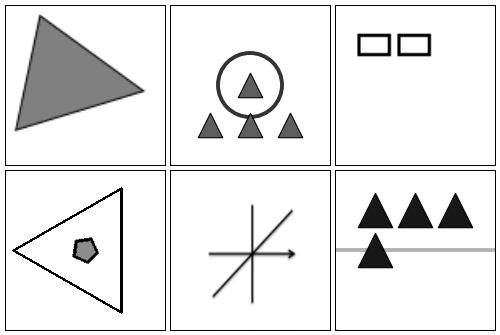}}
    ~
    ~
    \subfloat[Abstract shapes]{\includegraphics[width=.22\textwidth]{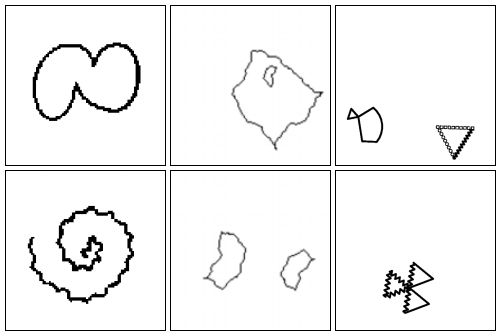}}
    \caption{
    \textbf{Input shapes.}
    AVR tasks can be composed of: (a) familiar and easy to comprehend \textit{geometric} shapes, that can be found in a wide suite of AVR problems; (b) novel---even randomly generated---\textit{abstract} shapes that rarely repeat across tasks.
    }
    \label{fig:input-shapes}
\end{figure}

As presented in Figs.~\ref{fig:human-avr-puzzle}-\ref{fig:arc},
AVR problems are composed of multiple panels, and each of them may comprise various shapes (see Fig.~\ref{fig:input-shapes} for some examples).
On a general note, we propose to differentiate between two types of shapes: \textit{geometric} and \textit{abstract}.
This distinction allows us to gain a clear perspective on:
1) solving which problems requires the ability to adapt to ever-changing inputs (abstract shapes);
2) for which tasks the knowledge of the finite vocabulary of shapes is sufficient when solving them (geometric shapes).

\subsubsection{Geometric shapes}
The prevalent number of AVR problems refer to matrices instantiated from a fixed set of geometric shapes that are rather easy to recognise and comprehend. 
This genre of problems includes Raven's Progressive Matrices, Visual Analogy Problems, Odd-one-out tasks, Visual Arithmetic Reasoning matrices, and Extrapolation problems.
Moreover, these shapes often appear in various variants and are characterised by certain properties.
For instance, the objects presented in Sandia matrices are defined by:
1) a shape: oval, rectangle, diamond, triangle, trapezoid, the letter T;
2) a variant: wide, narrow, tall, short;
3) an attribute: shading, orientation, size, count.
While such characteristics allow producing perceptually rich matrices, a solver with access to a finite vocabulary is generally able to fully comprehend the presented objects.

\subsubsection{Abstract shapes}
On the contrary, certain AVR tasks are composed of abstract shapes that come from a wide set of possibilities and rarely repeat across problems.
Such shapes underlie Bongard Problems and S-D tasks.
The use of a unique selection of atypical shapes leads to several consequences.
Firstly, it allows comparing human and machine visual reasoning performance in a fair setup, where both parties encounter the shapes for the first time.
Secondly, such a representation removes the need for complex perceptual modules that have to recognize various aspects of real-world images, such as texture or occlusion, that are often crucial for an accurate understanding of the scene. 

Consequently, this setting allows to fully focus on comprehension of the abstract relations present in the AVR tasks, and hence is especially well-suited for evaluating AVR capabilities.
We believe that abstract input setting should be a preferred mode in future AVR research.

\subsection{Hidden rules}
Even though AVR tasks with abstract shapes require the solving agent to be adaptable to novel inputs, the resultant AVR matrices, anyway, contain rather simple input shapes.
discovery~\cite{primi2001complexity}, which is further exacerbated if multiple rules are applied~\cite{carpenter1990one}.
Similar to the input shapes, AVR tasks can be divided into two main categories, depending on the type of the underlying rules.
We will refer to these categories as \textit{explicit} and \textit{abstract}, respectively.
Table~\ref{tab:hidden-rules} presents typical examples. 

\subsubsection{Explicit rules}
In the majority of AVR problems the rules come from a fixed and explicit set of well-defined possibilities.
Such explicit rules are often joint with geometric shapes.
Surprisingly, modern AVR benchmarks with explicit rules contain only a handful of unique relations, e.g. PGM contains 5 different rules: progression, XOR, OR, AND, consistent union; whereas RAVEN contains 4 such rules: constancy, progression, arithmetic, distribute three.
Even though the set of available rules is limited to only a few elements, its coupling with different shapes and attributes allows to generate huge suites of problem instances (e.g. PGM contains more than 1 million RPMs in each of its 8 regimes).
Quite surprisingly, AVR tasks generated from such seemingly small sets of rules often pose a challenge even for human solvers~\cite{zhang2019raven}.
\begin{table}
    \centering
    \caption{\textbf{Hidden rules.}
    AVR problems are centred around discovering hidden rules that govern shapes and their attributes present across images.
    Such relations can be either \textit{explicit}, where the rules are strict and well-defined, or \textit{abstract}, where the rules are described by either an abstract illustration or in natural language.
    The table provides examples for both cases.}
    \label{tab:hidden-rules}
    \begin{tabular}{p{0.14\textwidth}|p{0.26\textwidth}}
        \toprule
        \rowcolor{lightergray} \textbf{Explicit} & \textbf{Abstract} \\
        \midrule
        Logical operators: AND, OR & Each line connects two objects with the same shape but different size \\
        \midrule
        Constancy & Arrows are directed only horizontally \\
        \midrule
        Progression & Clockwise spirals have curly outlines \\
        \bottomrule
    \end{tabular}
\end{table}

\subsubsection{Abstract rules}
Contrary to AVR benchmarks with explicit rules, their counterparts with abstract relations are usually much more diverse.
Although abstract rules are often defined in less rigorous ways, they turned out to be really useful in determining the abstract reasoning skills of the test-taker.
As an example, both Bongard problems and same-different tasks employ abstract rules that define differences between two groups of panels.
This perspective places the solver in a never-seen-before scenario, which can be approached only by applying previously gained skills and knowledge in an entirely new setting.
Such problems lie at the heart of high-level intelligence~\cite{hofstadter1995fluid} and are more aligned with human IQ tests.
For the above reasons, we advocate for considering AVR problem setups with abstract rules as a more accurate approximation of human-level intelligence in machines.

\subsection{Target task}
In general, when solving AVR problems, the main difficulty for humans lies in discovering the abstract structure that governs the matrix.
After gaining this understanding, humans can solve AVR tasks irrespectively of whether the solution should be provided by selecting an answer from a fixed set of choices (\textit{classification}), recreating an element that matches the sequence (\textit{generation}), or expressing the answer in natural language (\textit{description}).
In contrast, ML/DL methods often struggle with solving certain target tasks, while excelling at others.

\subsubsection{Classification}
Classification---being the most often studied target task in the contemporary AVR literature---requires the solver to select an appropriate panel from the set of possible answers.
This task is offered in nearly all existing benchmarks, with the exception of the hand-crafted set of BPs and the matrices from DOPT.
In RPMs and VAPs, the test-taker is given a pre-defined set of possible answers, out of which only one correctly completes the matrix.
In BPs from Bongard-LOGO, the goal is to assign two test images to matching categories shown in left and right problem parts.
A similar approach is taken in the SVRT benchmark, which requires assigning a test image to one out of two categories.
Matrices from MNS also fall into this category, as the solver has to complete the matrix with an appropriate answer selected from a pre-defined set of choices, which in the case of MNS are integers ranging from 1 to 99.

Moreover, some problems offer an auxiliary target task consisting in recognition of additional meta-data related to a given problem instance.
As an example, besides completing the RPM task, the learning system may additionally explicitly predict the hidden rules that govern the matrix~\cite{barrett2018measuring,zhang2019raven,malkinski2020multilabel}.

\subsubsection{Generation}
Alternatively, instead of selecting an answer that correctly completes the matrix from a set of choices, a generative model may be considered that recreates the missing image (or part of it).
This setting is a default one in DOPT, where an image with a square, that follows a sequence, has to be generated, and in ARC, where an image that follows abstract rules expressed in few demonstrations needs to be generated.
In addition, baseline settings from some remaining AVR benchmarks can be adapted to form a generation task.
This idea was explored, for instance, in~\cite{pekar2020generating,zhang2021abstract} where the RAVEN dataset was tackled.
Instead of selecting RPM answers from a finite set of choices, the models had to predict a viable answer from scratch.

The above problem setting contains several challenges, as potentially many answers may be correct for the same problem instance due to possible variation in attributes not covered by the rules of the particular matrix.

Due to these challenges and limited availability of benchmarks created with the generation task in mind, we believe that additional sets of AVR problems are much required to facilitate development of successful AVR generative methods.

\subsubsection{Description}
The idea behind the initial set of hand-crafted BPs was to create a set of problems that represent abstract concepts which have to be discovered and described in natural language.
However, due to a limited number of problem instances, this formulation of a target task hasn't been considered to date in any of the proposed solutions.
While image captioning is a flourishing field with visible progress in recent years~\cite{bai2018survey,hossain2019comprehensive,stefanini2021show}, none of the existing methods tackles the problem of describing the abstract concepts present in AVR tasks in natural language.
Besides the mentioned set of hand-crafted BPs, there aren't any other benchmarks that could be used for evaluating the quality of image captioning methods in AVR settings. 

Definitely, developing such benchmark sets along with suitable methods for solving them remains an open problem and a challenging research path for future work.

\subsection{Cognitive function}
The entire domain of AVR problems is a rich testbed for two fundamental cognitive functions: \textit{completion} and \textit{discrimination}, which are highly relevant components of perceptual learning~\cite{casperson1950visual,gibson1955perceptual,bruner1957perceptual}.
In general, a successful AVR problem solver has to excel in both.

\subsubsection{Completion}
At the heart of many discussed AVR tasks lies the mechanism of \textit{completion}, where a selected sequence of images has to be completed such that all underlying patterns are preserved.
This genre of tasks includes RPMs, VAPs, matrices from MNS and both extrapolation settings from VAEC and DOPT.
Completion is often coupled with the target task of classification where an element that completes the sequence has to be chosen from a fixed set of provided answers, or sometimes with generation where the missing element has to be recreated from scratch.

\subsubsection{Discrimination}
A complementary cognitive function, that is required in practically all but MNS and DOPT problem instances, is centred around \textit{discrimination} between multiple possible answers.
For instance, when solving RPMs, a natural approach is to attempt to complete the matrix with each of the answer panels and then select the best-fitted one by means of their direct comparison.
The ability to discriminate is also required in BPs, $\mathrm{O}^3$, and S-D tasks, where a set of hidden rules that separate the provided panels has to be discovered.

Many works in cognitive literature have suggested that learning by discriminating positive from negative examples is of critical importance in the human learning process~\cite{gick1992contrasting,gick1992learning,gentner1994structural} and additionally promotes \textit{structural alignment}~\cite{gentner2001structural} that is beneficial in knowledge transfer~\cite{catrambone1989overcoming, bassok2003analogical}.

\subsection{Specific challenge}
AVR tasks pose a unique challenge on their own.
As illustrated by the already described dimensions of the taxonomy, these problems allow analysing capabilities of reasoning models from various perspectives.
Although, at first glance, they may seem like a set of puzzles with little relevance to other domains and more practical settings, in fact, quite the opposite is true.
AVR problems are often built around much more general challenges and the methods that excel in these tasks are likely to be revolutionary in other areas, as well.
Before we explore these connections in Section~\ref{sec:discussion}, we first introduce the main challenges posed by AVR problems.

\subsubsection{Domain transfer and knowledge generalisation}
In contemporary literature, AVR problems and related benchmark sets are often used to measure \textit{domain transfer abilities} of the tested methods in multiple ways:
(1) RAVEN contains matrices belonging to 7 different structural configurations, which can be used to verify whether the tested method is able to understand matrices with shapes arranged in novel ways~\cite{zhang2019raven,spratley2020closer};
(2) PGM defines 8 regimes that test model's ability to generalise to novel types of problem instances, e.g. to matrices with the same rules applied to novel objects and attributes, or to matrices with rules never encountered during the learning phase;
(3) VAPs from~\citet{hill2019learning} explicitly define problem instances where the solver has to identify the rule that governs the objects in a source domain and apply it to the objects in a (different) target domain.
Another example is DeepIQ~\cite{mandziuk2019deepiq}, a model trained to solve RPMs from the G-set, able to utilise this knowledge for solving Sandia RPMs and odd-one-out tasks.

Despite certain efforts, knowledge transfer between AVR problems remains a highly underexplored topic and so far no satisfactory solution has been proposed for this problem setting.

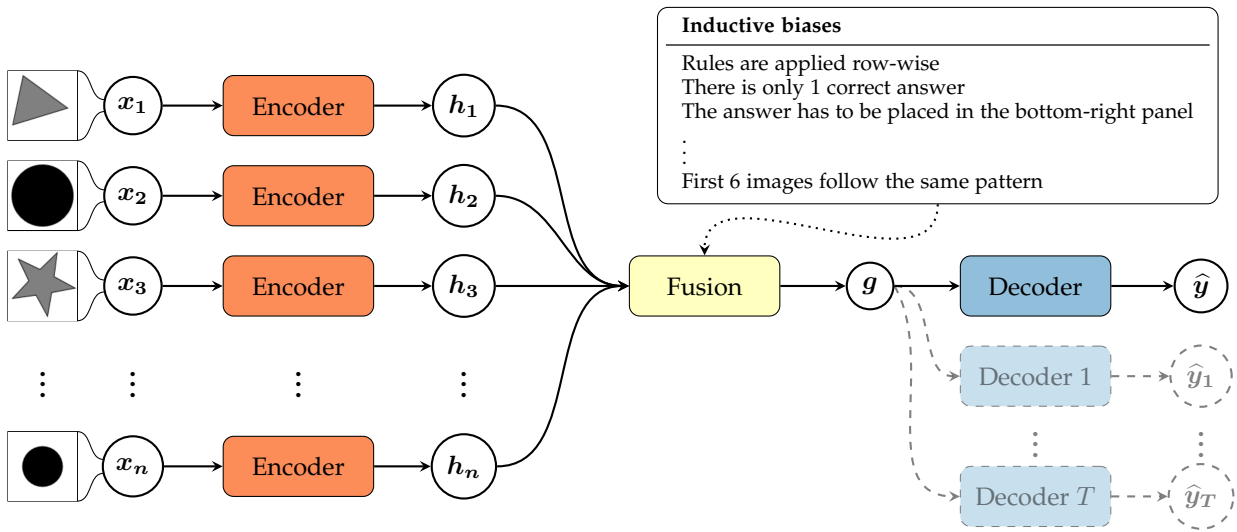
\begin{figure*}
    \tikzstyle{three sided panel} = [
        inner sep=0,
        draw=none,
        rectangle,
        append after command={
            \pgfextra{
                \begin{pgfinterruptpath}
                    \draw[draw,very thick] (\tikzlastnode.south east) edge ([xshift=-0.5\pgflinewidth]\tikzlastnode.south west);
                    \draw[draw,very thick] ([yshift=-0.5\pgflinewidth]\tikzlastnode.south west) edge ([yshift=0.5\pgflinewidth]\tikzlastnode.north west);
                    \draw[draw,very thick] ([xshift=-0.5\pgflinewidth]\tikzlastnode.north west) edge ([xshift=0.5\pgflinewidth]\tikzlastnode.north east);
                \end{pgfinterruptpath}
            }
        }
    ]
    \tikzstyle{panel} = [thick, rectangle, draw=black, inner sep=0]
    \tikzstyle{vector} = [thick, circle, draw=black, node contents={\bm{#1}}]
    \tikzstyle{optional vector} = [thick, circle, dashed, draw=black, opacity=0.5, node contents={\bm{#1}}]
    \tikzstyle{encoder} = [rectangle, rounded corners, minimum width=2cm, minimum height=0.8cm, text centered, draw=black, fill=myred]
    \tikzstyle{fusion} = [rectangle, rounded corners, minimum width=2cm, minimum height=0.8cm, text centered, draw=black, fill=myyellow]
    \tikzstyle{decoder} = [rectangle, rounded corners, minimum width=2cm, minimum height=0.8cm, text centered, draw=black, fill=myblue]
    \tikzstyle{auxdecoder} = [rectangle, dashed, rounded corners, minimum width=2cm, minimum height=0.8cm, text centered, draw=black, fill=myblue, opacity=0.5]
    \tikzstyle{arrow} = [thick,->,>=stealth]

    \begin{tikzpicture}
        \begin{scope}[node distance=1.2cm,on grid]
            \node (p1) [panel] {\includegraphics[width=.05\textwidth]{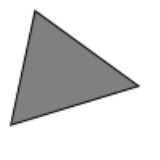}};
            \node (p2) [panel, below=of p1] {\includegraphics[width=.05\textwidth]{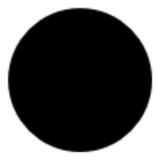}};
            \node (p3) [panel, below=of p2] {\includegraphics[width=.05\textwidth]{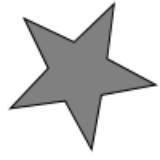}};
            \node (pi) [font=\Large, below=of p3] {\vdots};
            \node (pn) [panel, below=of pi] {\includegraphics[width=.05\textwidth]{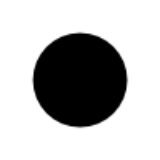}};

            \node (x1) [vector=$x_1$, right=of p1];
            \node (x2) [vector=$x_2$, below=of x1];
            \node (x3) [vector=$x_3$, below=of x2];
            \node (xi) [font=\Large, below=of x3] {\vdots};
            \node (xn) [vector=$x_n$, below=of xi];

            \node (e1) [encoder, right=of x1, xshift=1cm] {Encoder};
            \node (e2) [encoder, right=of x2, xshift=1cm] {Encoder};
            \node (e3) [encoder, right=of x3, xshift=1cm] {Encoder};
            \node (ei) [font=\Large, right=of xi, xshift=1cm] {\vdots};
            \node (en) [encoder, right=of xn, xshift=1cm] {Encoder};

            \node (h1) [vector=$h_1$, right=of e1, xshift=1cm];
            \node (h2) [vector=$h_2$, right=of e2, xshift=1cm];
            \node (h3) [vector=$h_3$, right=of e3, xshift=1cm];
            \node (hi) [font=\Large, below=of h3] {\vdots};
            \node (hn) [vector=$h_n$, right=of en, xshift=1cm];
            \node (ib) [rectangle, rounded corners, draw, right=of h1, xshift=5.1cm] {
                \footnotesize
                \begin{tabular}{l}
                    \textbf{Inductive biases} \\
                    \midrule
                    Rules are applied row-wise \\
                    There is only 1 correct answer \\
                    The answer has to be placed in the bottom-right panel \\
                    \vdots \\
                    First 6 images follow the same pattern \\
                \end{tabular}
            };
            \node (f) [fusion, right=of h3, xshift=2cm] {Fusion};
            \node (g) [vector=$g$, right=of f, xshift=1cm];
            \node (d) [decoder, right=of g, xshift=1cm] {Decoder};
            \node (y) [vector=$\widehat{y}$, right=of d, xshift=1cm];

            \node (d1) [auxdecoder, below=of d] {Decoder 1};
            \node (di) [font=\Large, opacity=0.5, below=of d1, yshift=0.4cm] {\vdots};
            \node (dn) [auxdecoder, below=of di, yshift=0.4cm] {Decoder $T$};
            \node (y1) [optional vector=$\widehat{y}_1$, right=of d1, xshift=1cm];
            \node (yi) [font=\Large, opacity=0.5, below=of y1, yshift=0.4cm] {\vdots};
            \node (yn) [optional vector=$\widehat{y}_T$, right=of dn, xshift=1cm];
        \end{scope}

        \draw (p1.north east) ++(-0.5\pgflinewidth,-0.5\pgflinewidth) to[out=0, in=180] ($ (x1.north west) + (-0.1, -0.2) $);
        \draw (p1.south east) ++(-0.5\pgflinewidth,0.5\pgflinewidth) to[out=0, in=180] ($ (x1.south west) + (-0.1, 0.2) $);

        \draw (p2.north east) ++(-0.5\pgflinewidth,-0.5\pgflinewidth) to[out=0, in=180] ($ (x2.north west) + (-0.1, -0.2) $);
        \draw (p2.south east) ++(-0.5\pgflinewidth,0.5\pgflinewidth) to[out=0, in=180] ($ (x2.south west) + (-0.1, 0.2) $);

        \draw (p3.north east) ++(-0.5\pgflinewidth,-0.5\pgflinewidth) to[out=0, in=180] ($ (x3.north west) + (-0.1, -0.2) $);
        \draw (p3.south east) ++(-0.5\pgflinewidth,0.5\pgflinewidth) to[out=0, in=180] ($ (x3.south west) + (-0.1, 0.2) $);

        \draw (pn.north east) ++(-0.5\pgflinewidth,-0.5\pgflinewidth) to[out=0, in=180] ($ (xn.north west) + (-0.1, -0.2) $);
        \draw (pn.south east) ++(-0.5\pgflinewidth,0.5\pgflinewidth) to[out=0, in=180] ($ (xn.south west) + (-0.1, 0.2) $);

        \draw [arrow] (x1) -- (e1); \draw [arrow] (e1) -- (h1);
        \draw [arrow] (x2) -- (e2); \draw [arrow] (e2) -- (h2);
        \draw [arrow] (x3) -- (e3); \draw [arrow] (e3) -- (h3);
        \draw [arrow] (xn) -- (en); \draw [arrow] (en) -- (hn);

        \draw [arrow] (h1.east) to[out=0, in=180] (f.west);
        \draw [arrow] (h2.east) to[out=0, in=180] (f.west);
        \draw [arrow] (h3.east) -- (f.west);
        \draw [arrow] (hn.east) to[out=0, in=180] (f.west);

        \draw [arrow,dotted] (ib.south) to[out=270, in=90] (f.north);

        \draw [arrow] (f.east) -- (g.west);
        \draw [arrow] (g.east) -- (d.west);
        \draw [arrow, dashed, opacity=0.5] (g.east) to[out=0, in=180] (d1.west);
        \draw [arrow, dashed, opacity=0.5] (g.east) to[out=-60, in=180] (dn.west);
        \draw [arrow] (d.east) -- (y.west);
        \draw [arrow, dashed, opacity=0.5] (d1.east) -- (y1.west);
        \draw [arrow, dashed, opacity=0.5] (dn.east) -- (yn.west);

    \end{tikzpicture}
    \caption{
    \textbf{Unified AVR model.}
    Existing DL models for solving AVR tasks can be roughly decomposed into 3 parts:
    \textit{encoder} -- that forms a representation of raw matrix images;
    \textit{fusion} -- which gathers information from panel representations into a single hidden representation;
    \textit{decoder} -- transforming the joint representation into the model's prediction.
    While this general approach is taken in the majority of existing solutions, the solution systems mainly differ in the inductive biases that are embedded in the model's architecture.
    Oftentimes, the biases form a strong assumption about the task being solved and are tailored towards a particular problem, which simplifies the chosen task, but makes the model less applicable to other related problems, where the embedded inductive biases are no longer relevant.  
    $\hat{y}$ represents the target task and ${\hat{y}}_1,\ldots,{\hat{y}}_T$ refer to (optional) auxiliary tasks.
    }
    \label{fig:unified-avr-model}
\end{figure*}

\subsubsection{Extrapolation}
A problem directly related to domain transfer is \textit{extrapolation}, where a model trained to operate on a fixed set of values of certain attributes is evaluated on problems with novel values of these attributes.
Such distinct generalisation regimes appear in:
(1) PGM, where train matrices contain shapes with lighter colour and smaller size, while test instances comprise of shapes with darker colour and bigger size;
(2) VAPs from~\citet{hill2019learning}, which borrow the concepts from PGM's extrapolation regime and additionally extend the number of sides a shape can have and the quantity of these shapes;
(3) Bongard-LOGO, where test matrices contain shape patterns generated by action programs with one additional instruction compared to the training ones, resulting in shapes with larger diversity.

What is more, both VAEC and DOPT directly focus on evaluating extrapolation by presenting sequences of squares with changing position and size.
Each such sequence has to be extended with an additional image that follows the pattern.
The key difficulty lies in the data distribution -- train sequences contain squares of smaller size and model evaluation is performed on sequences of larger squares.

\subsubsection{Arithmetic reasoning}
Another distinct challenge posed by AVR tasks resides at the intersection of visual and arithmetic reasoning.
The topic of solving math problems with computational methods has its roots in early domain literature~\cite{bobrow1964natural,mukherjee2008review,moses2012macsyma} and has received high interest in recent ML works that consider problems with symbolic~\cite{zaremba2014learning,lample2020deep,li2021isarstep} or text~\cite{kushman2014learning,hosseini2014learning,saxton2019analysing} inputs.
Merely endowing ML models with arithmetic reasoning turned out to be a staggering challenge in non-trivial problem settings~\cite{saxton2019analysing}.
However, only recently has the MNS dataset combined arithmetic problems with visual puzzles.

So far, no automatic learning method has achieved satisfactory performance, and the domain has yet to wait for effective machine solutions.

\subsubsection{Challenge \textit{per se}}
In addition to the described challenges posed by some AVR tasks, all existing AVR benchmarks can be viewed as self-contained challenges.
In this perspective, AVR tasks can be used to identify promising model components.
A module that performs exceptionally well in solving a particular AVR problem, assuming the model doesn't exploit shortcuts and doesn't embed too narrow inductive biases, might be of interest in solving other problems that involve abstract or relational reasoning.

An additional benefit of having a diverse set of tasks is their usage in the context of multi-task learning (MTL)~\cite{caruana1997multitask} and knowledge transfer.
As often shown in the MTL literature, a system that learns to solve multiple tasks might learn more general representations not achievable by models trained solely on a single task.

Consequently, additional AVR benchmarks are relevant in this area, even if they do not clearly focus on the main challenges covered by the AVR taxonomy.

    \section{Unified view of AVR models}\label{sec:overview-of-avr-models}

Inspired by the importance of AVR tasks in the evaluation of human intelligence, many works embarked on the quest to measure the performance of learning systems in solving these problems.
Initial approaches included models that employed hand-crafted rules~\cite{evans1964heuristic,foundalis2006phaeaco}, structure-mapping theory~\cite{lovett2007analogy,lovett2008computational,lovett2010structure,lovett2012modeling}, or similarity-based methods~\cite{mcgreggor2010fractal,kunda2010taking,mcgreggor2011fractally,mcgreggor2011finding,kunda2012reasoning}.
The progress made in these seminal works is comprehensively summarised in~\cite{hernandez2016computer}.

Despite promising results of the first above-mentioned computational models for solving AVR problems, the majority of recent literature gravitates toward DL systems.
This is largely attributed to their impressive performance that in many tasks reaches or even exceeds human level~\cite{mandziuk2019deepiq,wu2020scattering}.
However, the majority of existing learning systems are geared towards solving a single chosen task and are rarely evaluated in other settings.
Consequently, many models are discussed in isolation, despite clear similarities between the approaches.
In what follows we present a unified perspective on recent DL models for solving diverse AVR tasks.
We hope that this fusion of various computational threads from the recent AVR literature will be helpful in paving the way to general learning systems (Artificial General Intelligence) capable of solving disparate AVR problems.

In the existing DL systems for solving AVR tasks, three general components can be identified:
1) \textit{encoder}: that defines the way of processing and representing input panels of the problem instance;
2) \textit{fusion}: which determines how image representations are aggregated to fuse information from different matrix parts;
3) \textit{decoder}: specifying how the model arrives at its answer based on the latent problem representation.

While in some cases it is unclear where exactly the boundaries between these components should be determined, this rough partitioning allows gaining a clear understanding of what do different models have in common and where do they differ.
The above unified view on AVR models is depicted in Fig.~\ref{fig:unified-avr-model}.

\subsection{Encoder}
In general, each AVR problem instance can be considered as a set of panels (images) $X = \{x_i\}_{i=1}^n$, where $n$ is a dataset specific number of separate images in each problem instance (e.g. $n=16$ for RPMs from Sandia shown in Fig.~\ref{fig:rpm-sandia} and $n=4$ for $\mathrm{O}^3$ tests from G1-set shown in Fig.~\ref{fig:odd-one-out-1}).

Encoder, being the first component in each DL-based model for solving AVR tasks, defines how a raw panel (image) $x_i$ is converted to a hidden representation $h_i$.
$h_i$ is later aggregated with embeddings $\{h_j\}_{j \neq i}$ of other matrix panels in the fusion layer, so it is crucial that $h_i$ preserves as much semantic context as possible about the specific image.

The most common encoding approach is to use a simple neural block that extracts image low-level features.
For this purpose, some approaches flatten the image to a 1D sequence of pixels and employ a multi-layer perceptron (MLP) to transform it into a latent representation $h \in \mathbb{R}^d$, where $d$ is the dimension of the representation vector~\cite{mandziuk2019deepiq}.
Instead of MLP, other works apply CNNs and then flatten their output to again arrive at $h \in \mathbb{R}^d$~\cite{wu2020scattering}.
Alternatively, an identity transformation may be applied so that the original image is passed as-is to the fusion layer.
This idea is realized in several hierarchical models~\cite{hu2021stratified,zhuo2021effective} which are described in the following section.

\subsection{Fusion}
The next fundamental component of AVR models merges representations of individual panels $\{h_i\}_{i=1}^n$ into a unified representation $g$.
The main goal of the fusion component is to reason about rules that span multiple matrix panels, in contrast to the encoder that extracts panel-specific features.

On the one hand, in the majority of AVR works, the fusion module consists of simple MLPs -- composed of linear layers with non-linear activation functions and normalisations. Examples include BatchNorm~\cite{ioffe2015batch}, LayerNorm~\cite{ba2016layer}, or Temporal Context Normalisation~\cite{webb2020learning} that proved effective in solving extrapolation tasks.

On the other hand, despite certain similarities and shared components, the way of fusing individual panel representations, actually, differentiates the AVR models most.
By injecting diverse inductive biases into the model's architecture, the models are constructed with a particular problem in mind.
Consequently, the initial assumptions about the target task exert a big impact on the range of applicability of these methods to AVR problems.

In what follows, we catalogue existing AVR approaches with respect to the most common inductive biases they are built on.

\subsubsection{Separate panels}
Common to all contemporary approaches is the fundamental way of representing AVR matrices.
In general, when AVR tasks are used in human psychometric tests, the matrix is provided as a %whole 
single image.
Instinctively, humans start solving the matrix by decomposing it into separate panels.
In contrast, current AVR benchmarks provide already decomposed matrices as sets of individual panels.
Consequently, all existing works rely on this assumption and the availability of already separated representations of individual images forms a basis for the majority of remaining inductive biases.

A common way of exploiting this pre-processed representation is to stack all the images on top of each other (depthwise) and pass the resultant 3D matrix through CNN or ResNet~\cite{he2016deep}.
This approach was evaluated on many AVR tasks~\cite{hoshen2017iq,barrett2018measuring,zhang2019raven,hill2019learning,zhang2020machine,hu2021stratified,benny2020scale} and although it usually does not lead to achieving state-of-the-art results, it forms a reliable baseline.

Despite wide popularity, the assumption that an AVR task is provided as a set of separate images is perhaps a major obstacle on the way towards universal systems, capable of solving diverse AVR tasks that may come in a variety of forms.

\definecolor{enc1}{HTML}{FC9B59}
\definecolor{enc2}{HTML}{FC8D59}
\definecolor{enc3}{HTML}{FC7F59}
\definecolor{mlp1}{HTML}{F6F6C8}
\definecolor{mlp2}{HTML}{FFFFBF}
\definecolor{mlp3}{HTML}{FFF6BF}
\definecolor{connection}{HTML}{8d59fc}

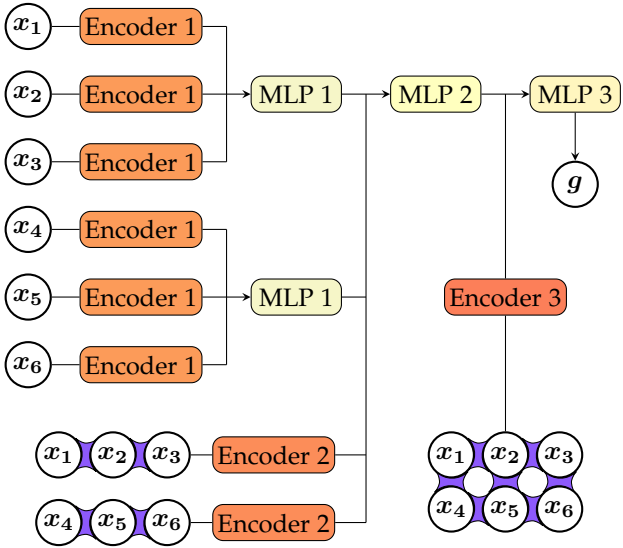
\begin{figure}
    \centering
    \tikzstyle{vector} = [thick, circle, draw=black, minimum width=\unit, minimum height=\unit, inner sep=0, node contents={\bm{#1}}]
    \tikzstyle{vect} = [thick, circle, draw=black, minimum width=\unit, minimum height=\unit, inner sep=0]
    \tikzstyle{emptymlp} = [rectangle, rounded corners, minimum width=3*\unit, minimum height=0.8*\unit, inner sep=0, text centered]
    \tikzstyle{mlp} = [rectangle, rounded corners, minimum width=2*\unit, minimum height=0.8*\unit, inner sep=0, text centered, draw=black, fill=mlp2]
    \tikzstyle{encoder1} = [rectangle, rounded corners, minimum width=2.7*\unit, minimum height=0.8*\unit, inner sep=0, text centered, draw=black, fill=enc1]
    \tikzstyle{encoder2} = [rectangle, rounded corners, minimum width=2.7*\unit, minimum height=0.8*\unit, inner sep=0, text centered, draw=black, fill=enc2]
    \tikzstyle{encoder3} = [rectangle, rounded corners, minimum width=2.7*\unit, minimum height=0.8*\unit, inner sep=0, text centered, draw=black, fill=enc3]
    \tikzstyle{arrow} = [->,>=stealth]

    \pgfmathsetmacro{\rbez}{0.04}
    \pgfmathsetmacro{\arcradius}{0.3}
    \newcommand \concat[3][]{
        \filldraw[#1] (#2.45)
            arc[start angle=45, end angle=-45, radius=\arcradius]
                .. controls +(45:\rbez) and +(180:\rbez) .. ($ (#2.-30) !0.5! (#3.210) $)
                .. controls +(0:\rbez) and +(135:\rbez) .. (#3.225)
            arc[start angle=225, end angle=135, radius=\arcradius]
                .. controls +(225:\rbez) and +(0:\rbez) .. ($ (#2.30) !0.5! (#3.150) $)
                .. controls +(180:\rbez) and +(-45:\rbez) .. cycle;
    }
    \newcommand \vconcat[3][]{
        \filldraw[#1] (#2.-45)
            arc[start angle=-45, end angle=-135, radius=\arcradius]
                .. controls +(315:\rbez) and +(90:\rbez) .. ($ (#2.-120) !0.5! (#3.120) $)
                .. controls +(270:\rbez) and +(45:\rbez) .. (#3.135)
            arc[start angle=135, end angle=45, radius=\arcradius]
                .. controls +(135:\rbez) and +(270:\rbez) .. ($ (#2.-60) !0.5! (#3.60) $)
                .. controls +(90:\rbez) and +(225:\rbez) .. cycle;
    }

    \begin{tikzpicture}
        \begin{scope}[node distance=1.5*\unit,on grid]
            \node (x1) [vector=$x_1$];
            \node (x2) [vector=$x_2$, below=of x1];
            \node (x3) [vector=$x_3$, below=of x2];
            \node (x4) [vector=$x_4$, below=of x3];
            \node (x5) [vector=$x_5$, below=of x4];
            \node (x6) [vector=$x_6$, below=of x5];
        \end{scope}

        \begin{scope}[node distance=1.5*\unit,on grid]
            \node (enc1) [encoder1, right=of x1, xshift=\unit] {Encoder 1};
            \node (enc2) [encoder1, right=of x2, xshift=\unit] {Encoder 1};
            \node (enc3) [encoder1, right=of x3, xshift=\unit] {Encoder 1};
            \node (enc4) [encoder1, right=of x4, xshift=\unit] {Encoder 1};
            \node (enc5) [encoder1, right=of x5, xshift=\unit] {Encoder 1};
            \node (enc6) [encoder1, right=of x6, xshift=\unit] {Encoder 1};
        \end{scope}

        \node (s1) [mlp, right=of enc2, xshift=-0.6*\unit, fill=mlp1] {MLP 1};
        \node (s10) at ($ (s1.west)!0.5!(enc2.east) $) {};
        \node (s11) at (s10 |- enc1) {};
        \node (s12) at (s10 |- enc3) {};
        \node (s2) [mlp, right=of enc5, xshift=-0.6*\unit, fill=mlp1] {MLP 1};
        \node (s20) at ($ (s2.west)!0.5!(enc5.east) $) {};
        \node (s21) at (s20 |- enc4) {};
        \node (s22) at (s20 |- enc6) {};

        \draw (x1.east) -- (enc1.west);
        \draw (enc1.east) -- (s11.center);
        \draw (s11.center) -- (s10.center);

        \draw (x2.east) -- (enc2.west);
        \draw [arrow] (enc2.east) -- (s1.west);

        \draw (x3.east) -- (enc3.west);
        \draw (enc3.east) -- (s12.center);
        \draw (s12.center) -- (s10.center);

        \draw (x4.east) -- (enc4.west);
        \draw (enc4.east) -- (s21.center);
        \draw (s21.center) -- (s20.center);

        \draw (x5.east) -- (enc5.west);
        \draw [arrow] (enc5.east) -- (s2.west);

        \draw (x6.east) -- (enc6.west);
        \draw (enc6.east) -- (s22.center);
        \draw (s22.center) -- (s20.center);

        % row hierarchy
        \begin{scope}[node distance=1.5*\unit,on grid]
            \node (enco1) [encoder2, below=of s2, xshift=-0.5*\unit, yshift=-2*\unit] {Encoder 2};
            \node (enco2) [encoder2, below=of enco1] {Encoder 2};
        \end{scope}

        \begin{scope}[node distance=1.5*\unit,on grid]
            \node (xx3) [vect] at ($ (enco1.west) + (-\unit, 0) $) {\bm{$x_3$}};
            \node (xx2) [vector=$x_2$, left=of xx3, xshift=0.3*\unit];
            \node (xx1) [vector=$x_1$, left=of xx2, xshift=0.3*\unit];
            \node (xx4) [vector=$x_4$, below=of xx1];
            \node (xx5) [vector=$x_5$, below=of xx2];
            \node (xx6) [vector=$x_6$, below=of xx3];
        \end{scope}

        \concat[fill=connection]{xx1}{xx2};
        \concat[fill=connection]{xx2}{xx3};
        \concat[fill=connection]{xx4}{xx5};
        \concat[fill=connection]{xx5}{xx6};

        \node (s3) [mlp, right=of s1, xshift=-0.6*\unit, fill=mlp2] {MLP 2};
        \node (s30) at ($ (s1)!0.5!(s3) $) {};
        \node (s31) at (s30 |- s2) {};
        \node (s32) at (s30 |- enco1) {};
        \node (s33) at (s30 |- enco2) {};

        \draw (xx3.east) -- (enco1.west);
        \draw (xx6.east) -- (enco2.west);
        \draw [arrow] (s1.east) -- (s3.west);
        \draw (s2.east) -- (s31.center);
        \draw (enco1.east) -- (s32.center);
        \draw (enco2.east) -- (s33.center);
        \draw (s33.center) -- (s30.center);

        \node (s4) [mlp, right=of s3, xshift=-0.6*\unit, fill=mlp3] {MLP 3};
        \node (s40) at ($ (s3)!0.5!(s4) $) {};
        \node (encod) [encoder3] at (s40 |- s2) {Encoder 3};
        \draw [arrow] (s3.east) -- (s4.west);
        \draw (encod.north) -- (s40.center);

        % eco hierarchy
        \begin{scope}[node distance=1.2*\unit,on grid]
            \node (xxx2) [vect] at (encod |- enco1) {\bm{$x_2$}};
            \node (xxx1) [vector=$x_1$, left=of xxx2];
            \node (xxx3) [vector=$x_3$, right=of xxx2];
            \node (xxx4) [vector=$x_4$, below=of xxx1];
            \node (xxx5) [vector=$x_5$, below=of xxx2];
            \node (xxx6) [vector=$x_6$, below=of xxx3];
        \end{scope}
        \draw (xxx2.north) -- (encod.south);

        \concat[fill=connection]{xxx1}{xxx2};
        \concat[fill=connection]{xxx2}{xxx3};
        \concat[fill=connection]{xxx4}{xxx5};
        \concat[fill=connection]{xxx5}{xxx6};
        \vconcat[fill=connection]{xxx1}{xxx4};
        \vconcat[fill=connection]{xxx2}{xxx5};
        \vconcat[fill=connection]{xxx3}{xxx6};

        \begin{scope}[node distance=1.5*\unit,on grid]
            \node (g) [vector=$g$, below=of s4, yshift=-0.5*\unit];
        \end{scope}
        \draw [arrow] (s4) -- (g);

    \end{tikzpicture}
    \caption{
    \textbf{Pre-defined hierarchies.}
    Stratified Rule Aware Network (SRAN)~\cite{hu2021stratified} that was designed with RPMs in mind, assumes a pre-defined hierarchy of matrix panels.
    Representations of single panels, rows, and pairs of rows are gradually combined to arrive at the final representation.
    The figure presents processing of the first two rows (1, 2) of the RPM -- analogous operation is applied to the pairs of rows (1, 3) and (2, 3).
    }
    \label{fig:sran}
\end{figure}

\subsubsection{Pre-defined hierarchies}
Thanks to the availability of a matrix pre-segmented into individual panels, many works incorporated a problem-specific arrangement of panels directly into the model's architecture.
Moreover, the underlying structure of AVR tasks is usually well-defined, which allows combining these individual panels in a task-relevant manner.
For instance, hidden rules in PGM matrices are only applied row-wise, which gives a clear hint into how representations of individual panels should be fused.

Based on these observations several task-specific architectures were proposed:
SRAN~\cite{hu2021stratified}, designed for solving RPMs, gradually aggregates the information from individual images, rows, and pairs of rows (see Fig.~\ref{fig:sran});
CoPINet~\cite{zhang2019learning} forms panel representations by adding together features along rows and columns;
LEN~\cite{zheng2019abstract} considers triples of objects only along rows and columns as inputs to a Relation Network~\cite{santoro2017simple};
DCNet~\cite{zhuo2021effective} explicitly creates representations of matrix rows and columns.

While all of these inductive biases turned out to be helpful in solving RPMs, the applicability of the resultant models to AVR tasks with other structures is limited.
The introduction of pre-defined hierarchies into model architectures is a common approach for improving results for a particular task, albeit at the cost of limiting the applicability to other problems.
To make the models suitable for solving diverse AVR tasks, pre-defined panel hierarchies could be replaced with automatic methods of discovering task-relevant panel combinations~\cite{wang2020abstract}.

\definecolor{mlp}{HTML}{F6F6C8}
\definecolor{mlp2}{HTML}{FFF6BF}
\definecolor{connection}{HTML}{8d59fc}

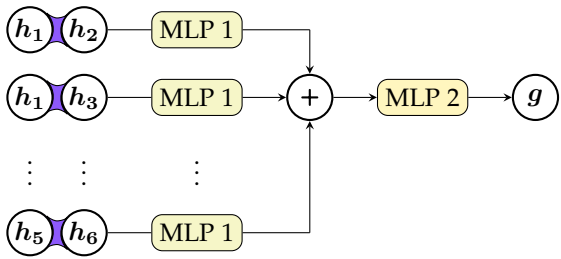
\begin{figure}
    \centering
    \tikzstyle{emptyvector} = [thick, circle, node contents={#1}, minimum width=1cm, minimum height=1cm, inner sep=0]
    \tikzstyle{vector} = [thick, circle, draw=black, minimum width=\unit, minimum height=\unit, inner sep=0, node contents={\bm{#1}}]
    \tikzstyle{vect} = [thick, circle, draw=black, minimum width=\unit, minimum height=\unit, inner sep=0]
    \tikzstyle{emptymlp} = [rectangle, rounded corners, minimum width=3*\unit, minimum height=0.8*\unit, inner sep=0, text centered]
    \tikzstyle{mlp} = [rectangle, rounded corners, minimum width=2*\unit, minimum height=0.8*\unit, inner sep=0, text centered, draw=black, fill=mlp]

    \tikzstyle{do path picture} = [path picture={
        \pgfpointdiff{\pgfpointanchor{path picture bounding box}{south west}}{
            \pgfpointanchor{path picture bounding box}{north east}
        }
        \pgfgetlastxy\x\y
        \tikzset{x=\x/2,y=\y/2}
        #1
    }]
    \tikzstyle{plus} = [thick, circle, draw=black, minimum width=\unit, minimum height=\unit, inner sep=0, do path picture={
        \draw [line cap=round] (-1/3,0) -- (1/3,0) (0,-1/3) -- (0,1/3);
    }]
    \tikzstyle{arrow} = [->,>=stealth]

    \pgfmathsetmacro{\rbez}{0.04}
    \pgfmathsetmacro{\arcradius}{0.3}
    \newcommand \concat[3][]{
        \filldraw[#1] (#2.45)
            arc[start angle=45, end angle=-45, radius=\arcradius]
                .. controls +(45:\rbez) and +(180:\rbez) .. ($ (#2.-30) !0.5! (#3.210) $)
                .. controls +(0:\rbez) and +(135:\rbez) .. (#3.225)
            arc[start angle=225, end angle=135, radius=\arcradius]
                .. controls +(225:\rbez) and +(0:\rbez) .. ($ (#2.30) !0.5! (#3.150) $)
                .. controls +(180:\rbez) and +(-45:\rbez) .. cycle;
    }
    \newcommand \vconcat[3][]{
        \filldraw[#1] (#2.-45)
            arc[start angle=-45, end angle=-135, radius=\arcradius]
                .. controls +(315:\rbez) and +(90:\rbez) .. ($ (#2.-120) !0.5! (#3.120) $)
                .. controls +(270:\rbez) and +(45:\rbez) .. (#3.135)
            arc[start angle=135, end angle=45, radius=\arcradius]
                .. controls +(135:\rbez) and +(270:\rbez) .. ($ (#2.-60) !0.5! (#3.60) $)
                .. controls +(90:\rbez) and +(225:\rbez) .. cycle;
    }

    \begin{tikzpicture}
        \begin{scope}[node distance=1.5*\unit, on grid]
            \node (p11) [vector=$h_1$];
            \node (p12) [vector=$h_2$, right=of p11, xshift=-0.3*\unit];

            \node (p21) [vector=$h_1$, below=of p11];
            \node (p22) [vector=$h_3$, below=of p12];

            \node (pi1) [emptyvector=\vdots, below=of p21];
            \node (pi2) [emptyvector=\vdots, below=of p22];

            \node (pn1) [vector=$h_5$, below=of pi1];
            \node (pn2) [vector=$h_6$, below=of pi2];
        \end{scope}

        \begin{scope}[node distance=1.5*\unit,on grid]
            \node (mlp1) [mlp, right=of p12, xshift=\unit] {MLP 1};
            \node (mlp2) [mlp, below=of mlp1] {MLP 1};
            \node (mlpi) [emptymlp, below=of mlp2] {\vdots};
            \node (mlpn) [mlp, below=of mlpi] {MLP 1};

            \node (s) [plus, right=of mlp2, xshift=\unit] {};
            \node (mlpp) [mlp, right=of s, xshift=\unit, fill=mlp2] {MLP 2};
            \node (g) [vector=$g$, right=of mlpp, xshift=\unit] {};
        \end{scope}
        \node (s1) at (s |- mlp1) {};
        \node (s2) at (s |- mlpn) {};

        \draw (p12) -- (mlp1);
        \draw (p22) -- (mlp2);
        \draw (pn2) -- (mlpn);

        \draw (mlp1) -- (s1.center);
        \draw [arrow] (s1.center) -- (s);

        \draw [arrow] (mlp2) -- (s);

        \draw (mlpn) -- (s2.center);
        \draw [arrow] (s2.center) -- (s);

        \draw [arrow] (s) -- (mlpp);
        \draw [arrow] (mlpp) -- (g);

        \concat[fill=connection]{p11}{p12};
        \concat[fill=connection]{p21}{p22};
        \concat[fill=connection]{pn1}{pn2};

    \end{tikzpicture}
    \caption{
    \textbf{Pair-wise relations.}
    Relation Network (RN)~\cite{santoro2017simple} operates on a set of $n$ objects ($n=6$ in the illustration), such as AVR panel embeddings $\{h_i\}_{i=1}^6$.
    The objects are combined into pairs $h_i \circ h_j$ by vector concatenation and processed by a single MLP that extracts pair-wise features.
    Next, a sum of these features is computed and passed through another MLP, that produces the final representation $g$ describing the whole set of input objects.
    }
    \label{fig:relation-network}
\end{figure}

\subsubsection{Pair-wise relations}
Another way of efficient processing of AVR matrices, enabled by the segmentation into individual panels, is to analyse relations between pairs of panel representations.
By comparing pairs of panels a model may, for instance, be trained to efficiently identify differences between them~\cite{mandziuk2019deepiq}.
Most notably, a pair-oriented approach was taken in the WReN model~\cite{barrett2018measuring,steenbrugge2018improving} that employs a Relation Network (RN)~\cite{santoro2017simple} depicted in Fig.~\ref{fig:relation-network}, which operates on a set of objects.
In the case of WReN, each object corresponds to a CNN embedding of an individual panel concatenated with the panel's absolute position in the matrix, expressed as a one-hot vector.
The embeddings are concatenated into pairs, each pair is processed by the same MLP, outputs are added together and finally processed by another MLP.

In contrast to the pre-defined hierarchies that are problem-specific, WReN---that considers pair-wise relations---was proven widely applicable to various AVR tasks~\cite{barrett2018measuring,zhang2019raven,hill2019learning,zhang2020machine,nie2020bongard,hu2021stratified,benny2020scale}.
Related pair-wise modules~\cite{vaswani2017attention,li2019area,shanahan2020explicitly} were found to be of crucial importance in other domains, as well.
Specifically, attention-based models can operate on a set of objects represented as a weighted sum of input features that do not necessarily have to correspond to separate input panels.
Generally speaking, unlike modules with pre-defined hierarchies, modules that consider pair-wise relationships are more flexible in solving different tasks.

\subsubsection{Single-choice tasks}
Further examples of how matrix pre-segmentation into individual panels could be exploited can be observed in the contemporary classification models for single-choice AVR tasks.
A typical design choice made in a wide suite of AVR models is to consider the set of matrix panels $\{x_i\}_{i=1}^n$ (also  referred to as the context panels) completed by each of the answer panels $\{a_j\}_{j=1}^m$ (see Fig.~\ref{fig:single-choice}).
Given the resultant set of panels $\{x_i\}_{i=1}^n \cup \{a_j\}$, the model can measure how aligned is the considered answer $a_j$ with the matrix context and express this alignment as a scalar score $s_j$.
With the help of the softmax function $\sigma$, the array of scores may be transformed into a probability distribution $\widehat{p} = \sigma([s_1, \ldots, s_m])$ and the answer corresponding to the highest probability will be chosen as the model's answer: $\widehat{y} = \argmax_j \{\widehat{p}_j\}_{j=1}^m$.

This approach is taken in the Wild-ResNet model~\cite{barrett2018measuring} that stacks the resultant $n + 1$ panels $\{x_i\}_{i=1}^n \cup \{a_j\}$ on top of each other and passes the matrix through ResNet to obtain a single scalar assessment.
Wild-ResNet was applied to solve AVR tasks from several benchmarks, including RPMs~\cite{barrett2018measuring,zhang2019raven,hu2021stratified,benny2020scale} and VAPs~\cite{hill2019learning}.

\definecolor{encoder}{HTML}{FC8D59}
\definecolor{fusion}{HTML}{FFFFBF}
\definecolor{decoder}{HTML}{91BFDB}

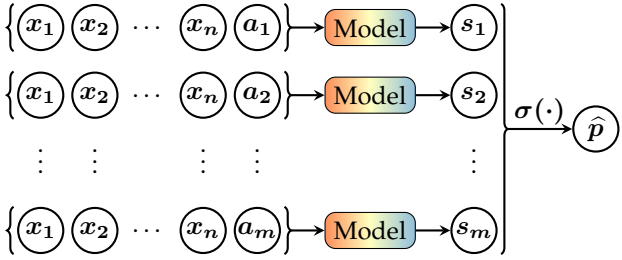
\begin{figure}

    \centering

    \tikzstyle{ecell} = [thick, circle, draw=black, minimum width=\unit, minimum height=\unit, inner sep=0, align=center]
    \tikzstyle{emptycell} = [thick, circle, minimum width=\unit, minimum height=\unit, inner sep=0, node contents={}]
    \tikzstyle{cell} = [emptycell, align=center, node contents={#1}]
    \tikzstyle{vector} = [cell, draw=black, node contents={\bm{#1}}]
    \tikzstyle{ldots} = [emptycell, node contents={\ldots}]
    \tikzstyle{vdots} = [emptycell, node contents={\rvdots}]
    \tikzstyle{emptymodel} = [rectangle, rounded corners, minimum width=2*\unit, minimum height=0.8*\unit, inner sep=0, node contents={}]
    \tikzstyle{model} = [emptymodel, align=center, draw=black, node contents=Model, shade, left color=encoder, right color=decoder, middle color=fusion]
    \tikzstyle{arrow} = [->,>=stealth,thick]

    \begin{tikzpicture}
        \begin{scope}[node distance=1.5*\unit, on grid]
            \node (x11) [vector=$x_1$];
            \node (x12) [vector=$x_2$, right=of x11, xshift=-0.3*\unit];
            \node (x1i) [ldots, right=of x12, xshift=-0.3*\unit];
            \node (x1n) [vector=$x_n$, right=of x1i, xshift=-0.3*\unit];
            \node (a1) [vector=$a_1$, right=of x1n, xshift=-0.3*\unit];

            \node (x21) [vector=$x_1$, below=of x11];
            \node (x22) [vector=$x_2$, right=of x21, xshift=-0.3*\unit];
            \node (x2i) [ldots, right=of x22, xshift=-0.3*\unit];
            \node (x2n) [vector=$x_n$, right=of x2i, xshift=-0.3*\unit];
            \node (a2) [vector=$a_2$, right=of x2n, xshift=-0.3*\unit];

            \node (xi1) [vdots, below=of x21];
            \node (xi2) [vdots, right=of xi1, xshift=-0.3*\unit];
            \node (xii) [emptycell, right=of xi2, xshift=-0.3*\unit];
            \node (xin) [vdots, right=of xii, xshift=-0.3*\unit];
            \node (ai) [vdots, right=of xin, xshift=-0.3*\unit];

            \node (xm1) [vector=$x_1$, below=of xi1];
            \node (xm2) [vector=$x_2$, right=of xm1, xshift=-0.3*\unit];
            \node (xmi) [ldots, right=of xm2, xshift=-0.3*\unit];
            \node (xmn) [vector=$x_n$, right=of xmi, xshift=-0.3*\unit];
            \node (am) [vector=$a_m$, right=of xmn, xshift=-0.3*\unit];

            \node (m1) [model, right=of a1, xshift=1*\unit];
            \node (m2) [model, right=of a2, xshift=1*\unit];
            \node (mi) [emptymodel, right=of ai, xshift=1*\unit];
            \node (mm) [model, right=of am, xshift=1*\unit];

            \node (s1) [vector=$s_1$, right=of m1, xshift=0.8*\unit];
            \node (s2) [vector=$s_2$, right=of m2, xshift=0.8*\unit];
            \node (si) [vdots, right=of mi, xshift=0.8*\unit];
            \node (sm) [vector=$s_m$, right=of mm, xshift=0.8*\unit];

            \node (p) [ecell] at ($ (s1)!0.5!(sm) + (2.7*\unit, 0) $) {$\bm{\widehat{p}}$};
        \end{scope}

        \draw [decoration={brace, mirror, raise=0.6*\unit}, decorate, thick] (x11.north) -- (x11.south);
        \draw [decoration={brace, mirror, raise=0.6*\unit}, decorate, thick] (x21.north) -- (x21.south);
        \draw [decoration={brace, mirror, raise=0.6*\unit}, decorate, thick] (xm1.north) -- (xm1.south);

        \draw [decoration={brace, raise=0.6*\unit}, decorate, thick] (a1.north) -- (a1.south);
        \draw [decoration={brace, raise=0.6*\unit}, decorate, thick] (a2.north) -- (a2.south);
        \draw [decoration={brace, raise=0.6*\unit}, decorate, thick] (am.north) -- (am.south);

        \node (b1) [inner sep=0] at ($ (a1.center) + (0.7*\unit, 0) $) {};
        \draw [arrow] (b1) -- (m1);
        \draw [arrow] (m1) -- (s1);

        \node (b2) [inner sep=0] at ($ (a2.center) + (0.7*\unit, 0) $) {};
        \draw [arrow] (b2) -- (m2);
        \draw [arrow] (m2) -- (s2);

        \node (bm) [inner sep=0] at ($ (am.center) + (0.7*\unit, 0) $) {};
        \draw [arrow] (bm) -- (mm);
        \draw [arrow] (mm) -- (sm);

        \draw [decoration={brace, raise=0.6*\unit}, decorate, thick] (s1.north) -- (sm.south);
        \node (bp) [inner sep=0] at ($ (s1)!0.5!(sm) + (0.7*\unit, 0) $) {};
        \draw [arrow] (bp) -- node [yshift=0.4*\unit] {$\bm{\sigma(\cdot)}$} (p);

    \end{tikzpicture}
    \caption{
    \textbf{Single-choice approach.}
    In single-choice tests with matrix (context) panels $\{x_i\}_{i=1}^n$ and answers $\{a_j\}_{j=1}^m$, a common modelling approach is to combine each answer separately with the set of context panels to arrive at $\{x_i\}_{i=1}^n \cup \{a_j\}$.
    For such a set of panels, the model can generate a score $s_j$ that defines how well is answer $a_j$ aligned with the remaining panels.
    A probability distribution over the set of possible answers can then be computed with softmax function $\widehat{p} = \sigma([s_1, \ldots, s_m])$ and the answer with the highest probability will be considered as the predicted solution $\widehat{y} = \argmax_j \{\widehat{p}_j\}_{j=1}^m$.
    }
    \label{fig:single-choice}
\end{figure}

A related approach is taken in~\cite{nie2020bongard} to solve BPs from Bongard-LOGO, where images from the left part together with a test image are stacked on top of each other and passed through a ResNet.
Analogous operation is applied to images from the right part.
Based on these two representations a discriminatory module produces a score that determines which side the test image aligns with.

A similar scheme is employed in WReN, however, in contrast to Wild ResNet, WReN generates an embedding for each of the panels separately.
Having the individual embeddings, WReN performs $m$ parallel passes through the RN module, each time using embeddings of all context panels together with the embedding of a selected answer, which produces a single scalar for the considered answer.
% This approach is applied in~\cite{barrett2018measuring,zhang2019raven,hill2019learning,hu2021stratified,benny2020scale}.

Likewise, a WReN-based module is considered in~\cite{nie2020bongard} which processes each image independently and then merges their representations using an RN.

While this paradigm of completing the context with one of the answers and generating an alignment score was proven useful in many other works~\cite{hahne2019attention,wu2020scattering,wang2020abstract,jahrens2020solving,spratley2020closer,rahaman2021dynamic}, the above approaches are fundamentally limited to these single-choice tasks in which the matrix is pre-segmented in advance.
Going forward, it will be beneficial to add explicit segmentation modules~\cite{minaee2021image} capable of dividing the matrix into individual panels.

\subsection{Decoder}
The wide suite of emerging AVR problems involves a number of distinct prediction tasks.
In the proposed AVR taxonomy, we have roughly divided them into the following three categories: \textit{classification}, \textit{generation} and \textit{description}.
Each of these target tasks requires a unique decoding module capable of producing answers in the expected format.
In what follows, we summarise the fundamental approaches in all 3 classes and discuss the importance of the auxiliary tasks that are orthogonal to this categorisation.

\subsubsection{Classification}
The set of AVR classification tasks mainly includes the already discussed single-choice tests where the context matrix has to be completed with a missing panel (Fig.~\ref{fig:single-choice}).

In addition, the target task of classification is fundamental to $\mathrm{O}^3$ matrices, where an odd element has to be identified amongst the set of matrix panels.
To this end, in~\cite{mandziuk2019deepiq} a heuristics-based scoring module was proposed that analysed feature-based differences extracted by a neural network.
In contrast to already described approaches where an alignment score is computed for the matrix completed by one of the answers, the approach proposed in~\cite{mandziuk2019deepiq} takes a holistic view and considers the matrix as a whole.

Another unique type of a classification task is presented by matrices from MNS where a single integer value has to be predicted that correctly fits into the hidden expression.
In this case, the missing integer may be one of 99 available choices and computing an alignment score of each one of them is impractical.
As a result, the existing models to solve visual arithmetic problems include a decoder implemented as a linear layer with 99 output neurons and the softmax activation function.

\subsubsection{Generation}
Another type of prediction challenge, posed by some AVR tasks, is image generation~\cite{goodfellow2014generative,gui2021review,jabbar2021survey}.
Instead of selecting an answer from a pre-defined set of choices, the answer has to be recreated from scratch.
Though the task is mainly emphasised in DOPT and ARC benchmarks, some works~\cite{hua2020modeling,pekar2020generating,zhang2021abstract,shi2021raven} considered the problem of generating RPM answers.
Answer generation enables a supplementary way of interpreting model's predictions -- by looking at what the model considers as an answer, we can roughly determine whether the model recognises the underlying structures or rather merely relies on visual shortcuts.

A common way of approaching an AVR generation task is to construct an autoencoder~\cite{hinton2006reducing,kingma2013auto,higgins2016beta} that firstly reduces the input image into a condensed latent representation and then uses this representation to recreate the image.
To excel in this task, the latent representation has to contain semantically meaningful information about the image, that can additionally be helpful in the target task of classification.

In some approaches, the task of image generation is treated as an auxiliary task used just for pre-training the encoder~\cite{hoshen2017iq,steenbrugge2018improving,mandziuk2019deepiq,tomaszewska2022duel}, while in other works the decoder is trained in parallel with the classification branch~\cite{webb2020learning,kim2020few}.
The decoder may be implemented as a sequence of MLPs~\cite{mandziuk2019deepiq}, or as a stack of transposed convolutions~\cite{hoshen2017iq,steenbrugge2018improving,webb2020learning,tomaszewska2022duel}, in both cases with the sigmoid activation in the last layer.

\subsubsection{Description}
While the spectrum of available classification and generation tasks in the AVR literature is quite wide, tasks involving description have not received comparable attention.
Since the seminal work~\cite{bongard1968recognition} that introduced Bongard Problems where hidden rules have to be described in natural language, none of the subsequent approaches succeeded in this task.
Instead, the problem was rephrased into simpler settings where description is replaced with binary classification~\cite{kharagorgiev2018solving,nie2020bongard} (cf. Section~\ref{sec:bongard-problems}).

On the other hand, recent progress in image captioning~\cite{bai2018survey,hossain2019comprehensive,stefanini2021show} and natural language generation coupled with scene understanding~\cite{ghosh2019generating,ilinykh2019tell,wei2021integrating} suggests that current learning systems are, in principle, capable of generating descriptions in natural language to reasoning problems with visual input.
This, in turn, suggests that the lack of successful methods for describing answers to AVR tasks in natural language may arise not from the lack of capacity of the proposed models, but rather from the unavailability of appropriate datasets on which such models could be trained.

An attempt to combine abstract reasoning with natural language was taken in~\cite{acquaviva2021communicating}, where the authors enrich ARC with natural language descriptions prepared by human annotators.
The descriptions provide instructions about how the matrices could be solved, which resembles the fundamental challenge posed by BPs, where a description of abstract rules that differentiate left and right matrix parts has to be generated.

Still, further research placed at the intersection of AVR and natural language generation is needed to advance the current ML systems in this human-specific ability.

\subsubsection{Auxiliary tasks}
In addition to the already described target tasks, some works consider auxiliary training objectives that are helpful in training the models for a downstream task.
Such additional objectives are achieved by attaching another decoder heads in parallel to the main decoder, as depicted in Fig.~\ref{fig:unified-avr-model}.
While some already mentioned works consider the image generation task as a supplementary objective, another common approach is to train the model to predict the underlying rules~\cite{barrett2018measuring}.
To this end, the task's structure is encoded as a multi-hot~\cite{barrett2018measuring,zhang2019raven} or one-hot~\cite{malkinski2020multilabel} vector and an auxiliary classification head is attached to the model with sigmoid activation in the last layer.
This allows predicting the rules that characterise the considered matrix.

Another group of notable auxiliary tasks adapt the family of contrastive losses~\cite{gutmann2010noise,oord2018representation} into the AVR setting, which improves the model's ability to discriminate between correct and wrong answers~\cite{malkinski2020multilabel,kim2020few}.

In summary, auxiliary tasks provide yet another means for specifying inductive biases helpful in the considered task, that can bring the model's focus to the important features identified with expert knowledge.
Unlike inductive biases that are embedded directly in model's architecture, auxiliary objectives are taken into account only during training, which makes it easier to adapt the model to novel tasks.

    \section{Discussion}\label{sec:discussion}
The AVR domain has seen an increasing research interest in recent years.
As a result, a wide suite of benchmarks emerged and many methods were proposed to tackle them.
At the same time, this increased attention sparked a discussion whether the field moves in the right direction.
Some speculations have been made about the unfairness of comparing progress in AVR with human performance.
In other discussions, the purpose of training models explicitly to solve AVR tasks is put into question.

While we generally share the above concerns, we also believe that current AVR research step-by-step narrows the gap between human and machine abilities in this highly challenging area.
In what follows, we present an optimistic view of the AVR field and formulate some prospects for future research.

\subsection{Humans vs Machines}
Achieving, or even surpassing, human performance in diverse tasks is the long-standing goal of AI research.
However, with the setup of current AVR benchmarks, it is far from obvious how, or even if, a fair comparison between humans and ML approaches can be conducted.
In this perspective, some recent works discussed how AVR tasks could be made more aligned with human psychometric tests.

\subsubsection{Few-shot vs multiple-epoch training}
One of the main differentiators in measuring human and machine performance in solving AVR tasks is the training regime.
While humans use AVR tasks mainly for the purpose of IQ evaluation, without extensive learning of how to solve them, current DL systems have to process huge training corpora before achieving any abstract reasoning abilities.

In~\cite{chollet2019measure} the author concludes that such training scheme measures how well does the algorithm learn certain skills, while we should be interested in evaluating the system's ability in \emph{acquiring} new skills.
Similarly,~\cite{mitchell2021abstraction} postulates measuring the system's abstract reasoning ability in a few-shot learning setting, which so far no method succeeded in.

While the above observations and claims are definitely valid, we would argue that those AVR problems where domain transfer is the main challenge, still allow making a fair comparison between human and machine performance.
In addition to ARC that explicitly tackles these issues by providing a set of matrices with unique tasks, let us take the PGM dataset as an example.
It is clear that generalisation regimes other than \textit{neutral} test the system's ability to solve novel tasks (please refer to~\cite{barrett2018measuring} for a detailed description of PGM regimes).
In such settings, the model's performance can't be improved with additional training examples as shown by the failure of all existing systems in solving matrices from more demanding PGM regimes~\cite{malkinski2022deep}.

Though benchmarks with small training corpora are a good way of ensuring that a fair comparison is made, datasets that test generalisation to novel types of problems, not encountered during training, offer a supplementary surface for comparing human and machine abstract reasoning abilities.

\subsubsection{White-box vs black/grey-box solutions}
Besides the training scheme, another important issue for conducting a fair comparison is the nature of the considered tasks.
A common issue encountered in the evaluation of DL-based systems arises from their lack of interpretability.
In effect, some systems with seemingly impressive performance are often found to exploit shortcuts in the datasets which renders the evaluation of their abstract reasoning ability difficult~\cite{hu2021stratified}.
As pointed out in~\cite{mitchell2021abstraction}, qualitative analysis of model predictions in generative tasks forms one way of gaining more confidence that the learning system indeed grasps the task of interest instead of relying on shortcuts.
In addition, tasks that focus on description form yet another set where interpretation of the model's predictions could be verified more convincingly.

Referring to the taxonomy proposed in Section~\ref{sec:introduction} it seems fair to say that the results in tasks of \textit{generation} and \textit{description} are more trustworthy.
In contrast, \textit{classification} settings might require additional interpretation of how the model arrives at its solution to validate whether the system really understands the task rather than relies on shortcuts.

Going one step further, another challenging setup would be to enrich the AVR benchmarks with human-like annotations describing the hidden rules (in a similar spirit to~\cite{suhr2017corpus}).
Based on these annotations, the trained models could learn not only to choose the set of underlying rules from a fixed selection, but additionally describe them in natural language.

Development of such explainable learning systems (XAI) is by many considered to be a necessary step on the path towards human-level intelligence.

\subsubsection{General vs task-specific approaches}
One of the crucial points when comparing human and machine reasoning performance is the range of applicability of the learning systems.
Current methods are usually evaluated on particular benchmarks, while humans are known for their intrinsic ability to solve disparate tasks.
In order to move forward, multi-task settings need to be considered which better align with the measurements of human intelligence.
While constructing general learning systems capable of solving diverse tasks is the grand and yet distant challenge, a first step could be to consider similar tasks according to the dimensions identified in the AVR taxonomy.
For instance, designing methods for solving various tasks with familiar geometric shapes, or related explicit rules, might be an achievable intermediate goal.

In summary, we postulate shifting the focus of the community towards solving multiple, but perhaps at first similar, AVR tasks.

\subsection{Beyond AVR domain}
Despite some criticism discussed in the previous section, recent advances made in the AVR domain have great potential to impact various related areas.

\begin{itemize}
    \item \textit{Discrimination}, being one of the cognitive functions fundamental to solving AVR problems, is often found in many practical applications such as robotics~\cite{sinapov2010odd} or anomaly detection~\cite{smets2011odd,schubert2014local}.
Contrastive AVR mechanisms~\cite{zhang2019learning,malkinski2020multilabel} that induce discrimination capabilities can be of high importance in such tasks.
    \item The most popular model with roots in AVR---the Relation Network---has already been applied to a wide array of tasks, including 3D human pose estimation~\cite{park20183d}, semantic segmentation~\cite{mou2019relation}, action recognition~\cite{sun2018actor}, reinforcement learning~\cite{zambaldi2018deep}, or self-supervised learning of visual representations~\cite{patacchiola2020self}.
    \item The perspective taken in the $\mathrm{O}^3$ problem was proven useful for defining a general weakly-supervised framework for learning representations~\cite{mohammadi2020odd}.
    \item Visual Analogy Problems that focus on extrapolation facilitated the construction of a general normalisation technique that can potentially be applied to problems beyond the AVR field.
    \item In the spirit similar to the seminal multitask learning paper~\cite{caruana1997multitask}, AVR benchmark sets demonstrated the importance of auxiliary tasks for building more transferable representations~\cite{barrett2018measuring,malkinski2020multilabel}.
    \item Also, a recent work~\cite{rahaman2021dynamic} used the PGM dataset to evaluate generalisation capabilities of a general compositional model.
\end{itemize}

\begin{figure*}[t]
    \centering
    \subfloat[]{\includegraphics[width=.24\textwidth]{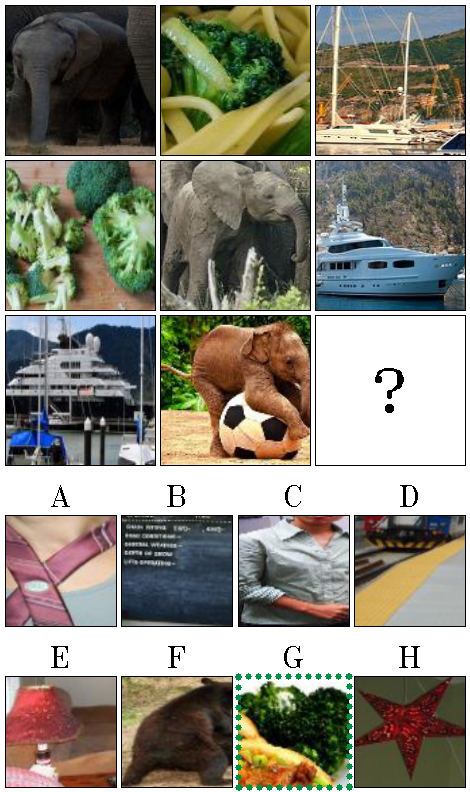}\label{fig:avr-inspired-vprom}}
    ~
    \subfloat[]{\includegraphics[width=.33\textwidth]{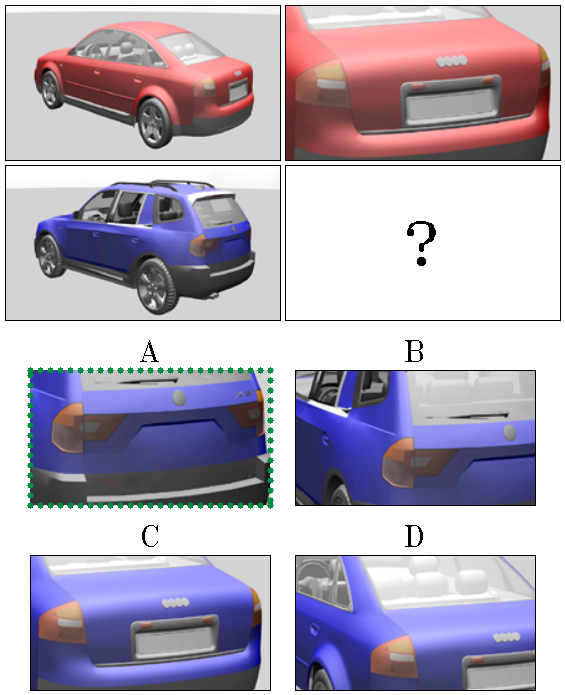}\label{fig:avr-inspired-cars}}
    ~
    \subfloat[]{\includegraphics[width=.37\textwidth]{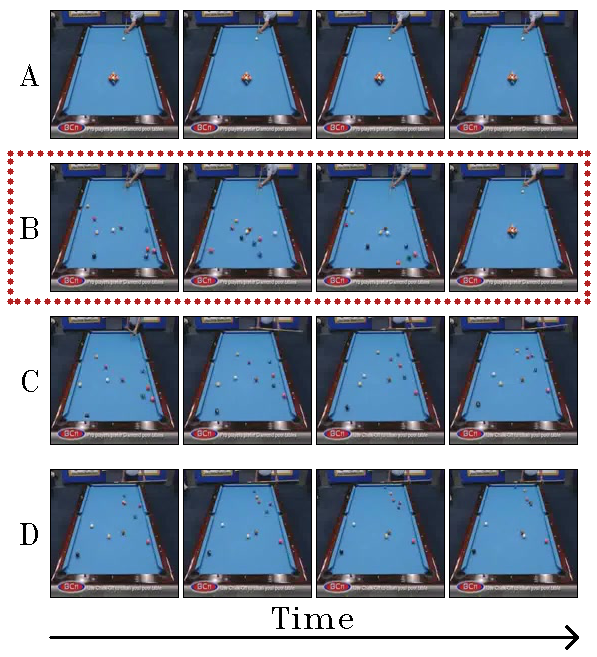}\label{fig:avr-inspired-ooo}}
    \caption{
    \textbf{AVR-inspired representation learning.}
    Problem formulations fundamental to AVR can be adopted as diagnostic tests for representation learning or act as pre-training tasks.
    (a) The V-PROM dataset~\cite{teney2020v} arranges real-world images selected from Visual Genome~\cite{krishna2017visual} into RPM structure. The trained model has to choose an answer that best satisfies the abstract rules defined by the context panels.
    (b) Images of 3D car models (or their fragments) from the ShapeNet dataset~\cite{chang2015shapenet} are arranged in an RPM-like grid. The trained model has to select an answer that best fits the analogy~\cite{ichien2021visual}.
    (c) Multiple subsequences are sampled from a video from UCF101~\cite{soomro2012ucf101} and the trained model has to detect the odd element -- a subsequence with perturbed temporal order of video frames~\cite{fernando2017self}.
    Correct answers are marked with a dotted green (for (a) and (b)) or red (for (c)) boundary.
    }
    \label{fig:avr-inspired-representation-learning}
\end{figure*}

On a general note, the specific challenges posed by AVR tasks identified in this survey, such as domain transfer, extrapolation, and arithmetic reasoning extend beyond AVR domain and are indispensable for developing human-level AI (or Artificial General Intelligence -- AGI).
AVR tasks already offer a wide suite of testbeds where initial attempts to develop particular skills of AGI agents can be made.

We are convinced that even if the performance of learning systems in AVR domain is still far from that of humans, or even if the performance gap between humans and machines cannot be assessed with current benchmarks, the underlying challenges posed by existing AVR datasets are of particular importance on the path to accomplishing human-level intelligence.

\subsection{AVR-like tasks for representation learning}
In addition to forming grounds for various generalisation challenges, AVR tasks serve as an inspiration for recent representation learning research, as illustrated in Fig.~\ref{fig:avr-inspired-representation-learning}.
In~\cite{teney2020v}, the authors propose the V-PROM dataset -- a collection of matrices with RPM structure where simple 2D shapes are replaced with real-life images (see Fig.~\ref{fig:avr-inspired-vprom}).
The matrices present fundamental rules such as counting or logical operators applied to pictures sampled from Visual Genome~\cite{krishna2017visual}.
The benchmark was created as a testbed for abstract high-level concept learning over real-life objects, which could not be measured with previous AVR datasets.

A new set of VAPs was also constructed in~\cite{ichien2021visual}.
Matrices from~\cite{ichien2021visual} focus on renderings of realistic cars from the ShapeNet dataset~\cite{chang2015shapenet}, as shown in Fig.~\ref{fig:avr-inspired-cars}.
The images vary in texture, shading and viewpoint.
Using this dataset, the authors presented a case where a general segmentation model performed better than task-specific architectures trained solely on the automotive VAPs~\cite{ichien2021visual}.

Yet another perspective was taken in~\cite{fernando2017self} where a self-supervised framework for video representation learning grounded in the $\mathrm{O}^3$ problem is proposed.
In this task, sequences of consecutive video frames are considered.
From each sequence, a few subsequences are extracted and one of them is perturbed by modifying the temporal order of frames.
As presented in Fig.~\ref{fig:avr-inspired-ooo}, the goal in this task is to identify such a perturbed (odd) subsequence.
Importantly, the training examples can be constructed automatically without human supervision.
Furthermore, the authors have shown that a network pre-trained on this self-supervised task learns meaningful representations of videos that are applicable to other downstream tasks such as action recognition on UCF101~\cite{soomro2012ucf101} and HMDB51~\cite{kuehne2011hmdb} datasets.

While both V-PROM and the dataset with car analogies were constructed as diagnostic tests, in the spirit of~\cite{fernando2017self} another possible application of both these sets could be their utilization as pre-training tasks for representation learning.
By focusing on high-level abstract concepts, the tasks might help inducing high-level general representations in visual models, which sets a promising avenue for future work.

Moreover, other datasets conceptually similar to V-PROM and automotive data, pertaining to other real-world domains are highly desirable in order to accelerate research at the intersection of AVR and practical settings.

\subsection{The main paths to move forward}
AVR domain has received special interest in recent years, due to the rapid progress in constructing DL models that excel in solving some of the tasks.
Based on these advances, several promising research directions can be identified.

\subsubsection{AVR benchmarks as self-contained challenges}
The AVR domain is rooted in the initial attempts of systematical evaluation of human intelligence.
While the settings of existing AVR benchmarks are far from those used in human studies, solving AVR tasks in their current form remains a research path of fundamental importance.
As already discussed, the AVR domain presents a ready-to-use set of problems that test generalisation abilities still not seen in current learning systems: \textit{flexibility}, by means of adaptability to novel domains, \textit{extrapolation}, or \textit{visual arithmetic reasoning}.

Further research on such methods, facilitated by the current benchmarks is one of the fundamental AVR research prospects.

\subsubsection{Towards Artificial General Intelligence}
Even though systems that excel in single AVR tasks are of high importance on their own, another path worth exploring is to consider AVR tasks with settings that correspond to human psychometric tests.
This way we can not only construct methods capable of possessing certain generalisation abilities, but additionally progress towards more human-like AI (AGI).

Despite impressive developments in recent years and attempts of reproducing human-like approaches to reasoning, learning and problem solving~\cite{hassabis2017neuroscience,lieto2018role}, existing learning systems are nowhere near human performance~\cite{lake2017building,marcus2020next}, and some even claim such levels are unattainable~\cite{fjelland2020general}.
While no clear path towards AGI seems to exist, the first step is to define how such progress can even be measured.
To this end, AVR tasks such as ARC~\cite{chollet2019measure} are crucial for measuring the machine progress towards truly intelligent systems.

Certainly, a lot of work is still ahead before the field would truly approach human-level performance in human-like AVR settings.

\subsubsection{Machine AI perspective in Human IQ research}
While ML research that assumes the above-mentioned alignment of ML experimental conditions with those used in human psychometric tests is undoubtedly one of the grand research avenues in AVR, there is another promising perspective for utilising ML systems that does not require such an alignment.

Recently, several attempts have been made to understand the organizational principles in human visual cortex by analysing how hierarchical DL models operate~\cite{yamins2014performance,kriegeskorte2015deep,yamins2016using}.
The researchers discovered that the intermediate representations learned by deep hierarchical CNNs correlate surprisingly well with processes found in biological brains~\cite{kriegeskorte2015deep,yamins2016using}, which brought new insights into a relatively poorly understood area of the visual cortex functioning~\cite{yamins2014performance}.

Per analogy, in parallel to bringing the human perspective into machine AVR studies, it might be a good time to investigate an opposite viewpoint, by bringing the perspective of machines into the human world.
This way we can gain another outlook for investigating human intelligence -- just as deep herarchical CNNs facilitated the formulation of novel insights about the visual cortex, analysis of representations learned by successful AVR machine solvers might help us unravel the mystery of impressive abstract visual reasoning skills in humans.

    \section{Conclusion}\label{sec:conclusion}
Abstract Visual Reasoning is a flourishing field with rapid progress in recent years.
To facilitate future well-informed research, in this work we have surveyed the emerging directions in AVR domain.

We have started with an introduction of the AVR taxonomy, which catalogues existing benchmark sets along 5 dimensions: input shapes, hidden rules, target task, cognitive function, and main challenge.
Next, based on this categorisation, we have reviewed the contemporary AVR problems.

Moreover, we have formulated a unified perspective on the introduced DL models for solving AVR tasks.
This view allows to distinguish general components relevant in many other settings from the problem-specific modules crafted with a particular dataset in mind.

Finally, we have discussed how a fair comparison between human and machine performance can be conducted, shed light on the bright side of AVR research, and discussed the connections between AVR and the representation learning literature.

Going forward, we have advocated for the role of AVR tasks in facilitating research on methods that exert specific generalisation abilities, as well as in accelerating advances towards human-level AI.

    \bibliographystyle{IEEEtranN}
    \bibliography{main}

\end{document}